\documentclass[aps,footinbib, notitlepage,superscriptaddress, longbibliography,eqsecnum]{revtex4-1}
\usepackage{graphicx}
 \usepackage{amssymb}
 \usepackage{amsmath}
 \usepackage{amsfonts}
\usepackage{overpic}
\usepackage{enumerate}
\usepackage{color}
\usepackage[dvipsnames]{xcolor}
\usepackage{xspace}
\usepackage[normalsize]{subfigure}
\usepackage{hyperref}
\usepackage{bm}
\usepackage{bbm}
\usepackage{empheq}
\usepackage[capitalize]{cleveref}

\DeclareFontFamily{OT1}{pzc}{}
\DeclareFontShape{OT1}{pzc}{m}{it}{<-> s * [1.10] pzcmi7t}{}
\DeclareMathAlphabet{\mathpzc}{OT1}{pzc}{m}{it}

\providecommand{\st}[1]{_{\text{#1}}}
\providecommand{\sfrac}[2]{#1/#2}

\def\onehalf{\frac{1}{2}}

\def\bra{\ensuremath{\langle}}
\def\ket{\ensuremath{\rangle}}
\def\eq{\st{eq}}

\def\const{\mathrm{const}}

\def\pd{\partial}
\def\tr{\mathrm{tr}}
\def\Imat{\mathbbm{1}}

\def\av{\bv{a}}

\def\uv{\bv{u}}
\def\ev{\bv{e}}

\def\Jcalv{\boldsymbol{\mathcal{J}}}

\def\yv{\bv{y}}
\def\zv{\bv{z}}
\def\Zv{\bv{Z}}
\def\vv{\bv{v}}

\def\xv{\bv{x}}
\def\wv{\bv{w}}

\def\Xm{\bv{X}}
\def\Ym{\bv{Y}}
\def\b0{\bv{0}}

\def\eff{\st{eff}}
\def\tot{\st{tot}}

\def\Acal{\mathcal{A}}
\def\Bcal{\mathcal{B}}

\def\Hcal{\mathcal{H}}
\def\Hc2{\Hcal^{(2)}}

\def\Jcal{\mathcal{J}}
\def\Ccal{\mathcal{C}}
\def\Dcal{\mathcal{D}}

\def\Lcal{\mathcal{L}}
\def\Mcal{\mathcal{M}}
\def\Ncal{\mathcal{N}}
\def\mcal{\mathpzc{m}}
\def\Ocal{\mathcal{O}}
\def\Pcal{\mathcal{P}}

\def\Rcal{\mathcal{R}}

\def\xib{\boldsymbol{\xi}}
\def\hyp13{{_1 F_3}}

\def\d{\mathrm{d}}
\def\cov{\mathrm{cov}}
\def\rank{\mathrm{rank}}
\def\diag{\mathrm{diag}}
\def\vect{\mathrm{vec}}
\def\var{\mathrm{var}}
\def\reals{\mathbb{R}}

\newcommand{\bitem}{\begin{itemize}}
\newcommand{\eitem}{\end{itemize}}
\newcommand{\benum}{\begin{enumerate}}
\newcommand{\eenum}{\end{enumerate}}
\newcommand{\btab}[1]{\begin{tabular}{#1}}
\newcommand{\etab}{\end{tabular}}
\newcommand{\beq}{\begin{equation}}
\newcommand{\eeq}{\end{equation}}
\newcommand{\beqn}{\begin{equation*}}
\newcommand{\eeqn}{\end{equation*}}

\newcommand{\bv}[1]{\mathbf{#1}}

\begin{document}
\title{Weight fluctuations in (deep) linear neural networks and a derivation of the inverse-variance flatness relation}
\author{Markus Gross}
\email{markus.gross@dlr.de}
\author{Arne P. Raulf}
\author{Christoph R\"ath}
\affiliation{Deutsches Zentrum f\"ur Luft- und Raumfahrt, Institute for AI Safety and Security, Sankt Augustin \& Ulm, Germany}
\date{\today}

\begin{abstract}
We investigate the stationary (late-time) training regime of single- and two-layer underparameterized linear neural networks within the continuum limit of stochastic gradient descent (SGD) for synthetic Gaussian data. In the case of a single-layer network in the weakly underparameterized regime, the spectrum of the noise covariance matrix deviates notably from the Hessian, which can be attributed to the broken detailed balance of SGD dynamics. The weight fluctuations are in this case generally anisotropic, but effectively experience an isotropic loss. 
For an underparameterized two-layer network, we describe the stochastic dynamics of the weights in each layer and analyze the associated stationary covariances. We identify the inter-layer coupling as a distinct source of anisotropy for the weight fluctuations. In contrast to the single-layer case, the weight fluctuations are effectively subject to an anisotropic loss, the flatness of which is inversely related to the fluctuation variance. We thereby provide an analytical derivation of the recently observed inverse variance-flatness relation in a model of a deep linear neural network.
\end{abstract}

\maketitle

\section{Introduction}

\subsubsection{Motivation} 
Deep neural networks have proven to be formidable tools in the area of machine learning \cite{goodfellow_deep_2016,aggarwal_neural_2018,murphy_probabilistic_2022}. Their ability to generalize from data often hinges on the dynamics of their training processes, notably the stochastic gradient descent (SGD) method. 
While the success of these networks in practical applications is well-documented, the underlying principles governing their learning dynamics, especially concerning weight fluctuations and loss landscapes, are still not fully understood. 
In order to address some of these aspects analytically, we focus here on one- and two-layer linear networks trained on synthetic Gaussian data with noisy labels. 
These simplified models have been proven to share many aspects of the learning dynamics of full nonlinear networks \cite{saxe_exact_2013, saxe_mathematical_2019, arora_implicit_2019, advani_highdimensional_2020, hastie_surprises_2022, braun_exact_2022}.
In fact, in the stationary regime at the end of training, a deep nonlinear network typically follows a linearized dynamics (multivariate Ornstein-Uhlenbeck process with correlated noise) \cite{mandt_stochastic_2017, jastrzebski_three_2018, chaudhari_stochastic_2017}.
We employ here the continuum description of SGD in terms of a Langevin equation, which allows us to analyse training dynamics and steady-state properties from a statistical physics perspective \cite{engel_statistical_2001,bahri_statistical_2020,huang_statistical_2021}.

One motivation of the present study stems from the recently observed inverse relation between the variance of a weight fluctuation mode and the flatness of the loss along the direction of that mode \cite{feng_inverse_2021}. While this inverse variance-flatness relation (IVFR) has been investigated from various perspectives \cite{feng_inverse_2021, xiong_stochastic_2022, yang_stochastic_2023, adhikari_machine_2023}, we seek here a first-principles derivation in a neural network model.
This requires, as a preparatory step, also a detailed understanding of the various sources of anisotropy of the weight fluctuations, which we address in the first part of this study, focusing on single-layer linear network.
Complementing recent works on this aspect \cite{kunin_limiting_2021,ziyin_strength_2021}, we consider in particular the slightly oversampled regime near the interpolation threshold (i.e., the point where the sample number equals the input dimension) and investigate the dependencies on the parameter count, sample number and label noise. 
We find that arguably the simplest model showing an IVFR is a two-layer linear network. In fact, the inter-layer coupling of the weights is a crucial ingredient responsible for the loss anisotropy, which therefore does not occur in the case of a single-layer network.
However, even for a single-layer network, the weight fluctuations are anisotropic except in the highly oversampled regime.
In a physical sense, the IVFR indicates that the effective noise temperature is higher in regions where the loss valleys are narrow, thus effectively biasing the model towards flat regions of the loss. Notably, wide loss valleys are typically associated with improved generalization ability of the network \cite{hochreiter_flat_1997,jastrzebski_three_2018,soudry_implicit_2018}.

Another motivation of our study stems from the observation that, while the (deterministic) gradient flow dynamics of deep linear networks has been addressed in several studies \cite{saxe_exact_2013,lampinen_analytic_2019,braun_exact_2022}, the corresponding layer-wise stochastic gradient flow dynamics has received less attention so far. 
A notable exception is the online learning in two-layer nets, which can be assessed analytically in certain regimes \cite{saad_online_1995, riegler_online_1995, goldt_dynamics_2019, arnaboldi_highdimensional_2023}. Recently, effects of stochasticity in two-layer linear nets have been analyzed to explain the observed bias towards sparse weight distributions \cite{chen_stochastic_2023}.
In the present work, we focus on supervised learning of mini-batches of data and study the associated stochastic dynamics of the weights in each layer within a Langevin approach.

\subsubsection{Related work}
\textit{Continuum modeling of SGD.}
The modeling of SGD in terms of a stochastic differential equation (SDE) -- in the sense that the former emerges from a Euler-Maruyama time-discretization of the latter -- is an established approach in the literature \cite{mandt_continuoustime_2015, mandt_variational_2016, mandt_stochastic_2017, li_stochastic_2017, hu_diffusion_2017, jastrzebski_three_2018, chaudhari_stochastic_2017, wojtowytsch_stochastic_2021a, wojtowytsch_stochastic_2021}.
In particular, the steady-state dynamics at the end of training can be described as a multivariate Ornstein-Uhlenbeck process \cite{mandt_stochastic_2017, jastrzebski_three_2018, ali_implicit_2020, kunin_limiting_2021, li_what_2021}.
Despite being formally derived in the limit of infinitely small learning rate \cite{yaida_fluctuationdissipation_2019}, the continuum approximation to SGD is typically found to be meaningful also for finite learning rates \cite{li_validity_2021} and is has been successfully used to study various aspects of the training of neural networks. 
The noise due to the mini-batch sampling of gradients in SGD plays an important role for the generalization ability of a model as it directs the weights towards wider minima of the loss, which improves generalization \cite{hochreiter_flat_1997, jastrzebski_three_2018, zhang_energyentropy_2018, smith_bayesian_2017, zhu_anisotropic_2018, nguyen_first_2019, xie_diffusion_2020, chen_anomalous_2020, mori_powerlaw_2022, wu_alignment_2022, yang_stochastic_2023, sclocchi_dissecting_2023} and helps escaping from saddle points and sharp minima \cite{hu_diffusion_2017, daneshmand_escaping_2018, xie_diffusion_2020, wang_theoretical_2023}.

For sufficiently large mini-batch sizes, the central limit theorem implies a Gaussian distribution of the gradient noise \cite{mandt_stochastic_2017, jastrzebski_three_2018, wu_revisiting_2021, wu_noisy_2019}, although deviations towards more heavy-tailed distributions have been observed \cite{simsekli_heavytailed_2019, simsekli_tailindex_2019, xie_rethinking_2022}.
Moreover, the gradient noise is assumed to be uncorrelated in time (see, however, Ref.\ \cite{kuhn_correlated_2023}).
A crucial property of the noise is its anisotropy, i.e., its dependence on the direction in parameter space \cite{zhu_anisotropic_2018, xie_diffusion_2020, ziyin_strength_2021, feng_inverse_2021, wang_theoretical_2023}.
In the simplest case, this anisotropy is modeled by taking its covariance to be proportional to the Hessian \cite{liu_noise_2021, kunin_limiting_2021, wu_revisiting_2021}. However, this approximation becomes exact only in the infinite sample limit (see discussion in \cref{sec_noise_covar} below). 
In fact, the discrepancy between noise covariance and Hessian implies that detailed balance is broken in the stationary state reached by SGD, giving rise to intriguing non-equilibrium behavior such as persistent currents \cite{chaudhari_stochastic_2017, kunin_limiting_2021, adhikari_machine_2023}.
In the context of linear regression, effects of parameter-dependent (multiplicative) noise have been addressed in \cite{ali_implicit_2020}.  
The relation of the noise covariance matrix to the Hessian and the Fisher matrix is further analyzed in \cite{thomas_interplay_2020}.

\textit{(Deep) Linear nets.}
Single-layer linear networks often admit closed form analytic solutions which qualitatively explain many properties observed in deep nonlinear networks (see, e.g., \cite{krogh_generalization_1992, engel_statistical_2001, advani_highdimensional_2020, mei_generalization_2019, hastie_surprises_2022}). 
In the limit of large width, the neural tangent kernel provides a linear approximation of a nonlinear network around initialization \cite{lee_wide_2019, jacot_neural_2018}.
Moreover, the steady-state dynamics of a trained nonlinear network becomes effectively linear -- a fact that, e.g., is utilized in the construction of Bayesian sampling methods \cite{mandt_stochastic_2017, chaudhari_stochastic_2017}.
Notably, linear networks with two or more layers have a nonlinear gradient descent dynamics of the weights and feature a non-convex loss landscape \cite{saxe_exact_2013, laurent_deep_2018, achour_global_2021}.
Deep linear nets have emerged as important toy models for the study of more general networks with nonlinear activation functions, enabling fundamental insights into training dynamics, generalization ability, implicit regularisation, and spectral bias \cite{saxe_exact_2013, lampinen_analytic_2019, arora_optimization_2018, ji_gradient_2018, du_width_2019, arora_implicit_2019, gidel_implicit_2019, saxe_mathematical_2019, tarmoun_understanding_2021, braun_exact_2022, chizat_infinitewidth_2022, chen_stochastic_2023}. 

\textit{Inverse variance-flatness relation (IVFR).}
The inverse relation between the magnitude of the steady-state weight fluctuations and the flatness of the loss has been first discussed in \cite{feng_inverse_2021}.
As emphasized in \cite{xiong_stochastic_2022}, using the formalism of \cite{ao_potential_2004, kwon_structure_2005, ao_stochastic_2006}, the actual effective potential (understood as the inverse of the covariance matrix of the weight distribution) is not the loss, but the product of the inverse of the (generally anisotropic) diffusion term and the loss. This indeed follows also from the classic theory of stochastic processes \cite{risken_fokkerplanck_1989, gardiner_stochastic_2009, kunin_limiting_2021} and has been realized in various machine-learning contexts \cite{mandt_stochastic_2017,chaudhari_stochastic_2017}.
In Ref.\ \cite{yang_stochastic_2023}, IVFR has been derived within a simple dynamic model of slow and fast degrees of freedom.
The stability of IVFR under dropout has been discussed in \cite{zhang_dropout_2021}.

\subsubsection{Outline of the present work}
We first consider learning of synthetic data via SGD in a single-layer linear network (\cref{sec_linreg}). Employing the continuum description of the training dynamics, we show that the spectrum of the noise covariance matrix in the stationary state deviates significantly from the Hessian, which we relate to the well-known broken detailed balance of the SGD dynamics. As a consequence, the weight fluctuations in the underparameterized regime are generally anisotropic, i.e., their amplitude depends on the directions of the principle modes.

Proceeding to a two-layer linear network (\cref{sec_2LNet}), we determine a continuum description of the layer-wise training dynamics under SGD. We obtain coupled stochastic differential equations for the weights and determine expressions for the drift and diffusion matrices in the large batch regime. These matrices turn out to be generally rank deficient due to the coupling between the network layers. As a consequence the weight fluctuations in each layer are anisotropic. 

We finally apply our results to study the shape of the loss under a perturbation along a principle weight fluctuation mode. In the case of a single-layer network, the loss remains isotropic despite the anisotropy of the weight fluctuations. For a two-layer network, we confirm the recently observed \cite{feng_inverse_2021} inverse relation between the amplitude of a weight fluctuation mode and the flatness of the loss along the direction of that mode. The coupling between the layers of the network is crucial for our result. The present work thus provides a derivation of the inverse variance-flatness relation in an analytically tractable model of a deep neural network.

\section{Single-layer linear networks}
\label{sec_linreg}

In this section, we consider the fluctuating dynamics of the weights of an underparameterized single-layer linear network. The network evolves under a mean-square loss and thus solves a linear regression problem.
The present linear framework can be matched to a nonlinear network in a stationary state (see \cref{app_nonlinearNN}).

\begin{figure}[t]\centering
    {\includegraphics[height=0.22\linewidth]{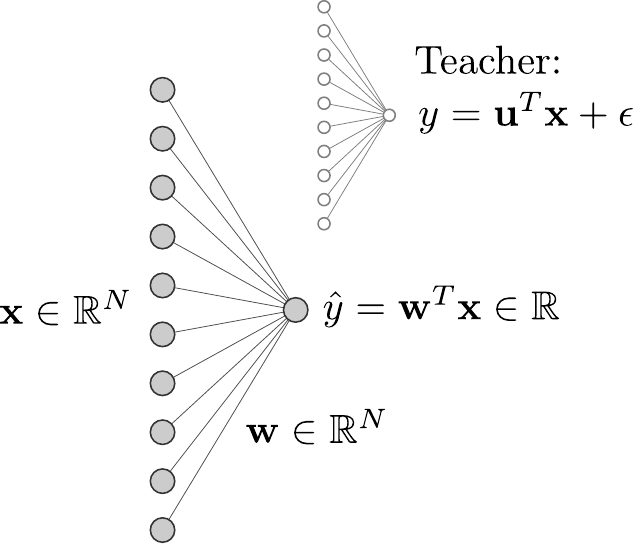} }
    \caption{Single-layer linear neural network model. The student network (left) transforms the input $\xv$ to the output $\hat y$ using its learned weights $\wv$, while the teacher network (right) generates the targets (``labels'') $y$ using fixed random weights $\uv$ and noise $\epsilon$ (see text).} 
    \label{fig_1lnet}
\end{figure}

\subsection{Linear regression}
\label{sec_statics}
We begin by recapitulating the theory of linear regression that is necessary for the subsequent parts of our study \cite{bishop_pattern_2006,hastie_elements_2009}. 
The training data consists of $P$ samples $\xv^\mu\in\reals^{N}$ ($\mu\in 1,\ldots, P$), the components being i.i.d.\ Gaussian random variables $x^\mu_i\sim \Ncal(\mu_x,\sigma_x^2)$ with mean $\mu_x$ and variance $\sigma_x^2 = 1/N$ \footnote{This variance is chosen in order to obtain quantities of $\Ocal(1)$ after applying random teacher weights as given in \cref{eq_teacher_y}.}. 
The targets $y^\mu\in \reals$ are provided by a linear teacher model:
\beq y^\mu = \uv^T \xv^\mu  + \epsilon^\mu,
\label{eq_teacher_y}\eeq 
with a fixed random weight vector $\uv\in\reals^{N}$ with $u_i\sim\Ncal(0,\sigma_u^2)$ (we typically take $\sigma_u^2=1$) and a residual error $\epsilon^\mu\sim \Ncal(0,\sigma_\epsilon^2)$ modeled as a i.i.d.\ Gaussian noise.
The error term describes the residual dependence between inputs and targets that can not be captured by a linear relationship. In a slight abuse of terminology, we will also in the present regression context refer to $y$ and $\epsilon$ as `label' and `label noise', respectively \cite{kunin_limiting_2021, ziyin_strength_2021}.
The complete set of the training data $(\xv^\mu, y^\mu)_{\mu\in\{1,\ldots,P\}}$ is called \emph{batch}, while a randomly chosen subset $(\xv^\mu, y^\mu)_{\mu\in \Bcal}$ of size $S<P$ with $\Bcal=\{\pi(1),\ldots, \pi(S)\}$ (and $\pi$ being a random permutation of the $P$ elements) is called a \emph{mini-batch}.
The case of statistically independent random labels $y^\mu$, which occasionally is useful in order to analyze fluctuations, corresponds to the replacements $\uv=0$, $\epsilon^\mu \to y^\mu $. 
We define a design matrix $X\in\reals^{N\times P}$ with the columns representing the samples $\xv^\mu$, and, correspondingly, a noise vector $\boldsymbol{\varepsilon}=(\epsilon^1,\epsilon^2,\ldots,\epsilon^P)\in\reals^{1\times P}$, such that \cref{eq_teacher_y} can be written as 
\beq Y = \uv^T X + \boldsymbol{\varepsilon} \in\reals^{1\times P}.
\eeq 
The linear student network is defined by $y= \wv^T \xv$, with the learnable weights $\wv \in\reals^{N}$ (see \cref{fig_1lnet}). 
The full-batch loss takes the form of a least-squares regression model:
\beq L(\wv) =\frac{1}{2 P} \| \wv^T X  - Y \|^2 = \frac{1}{2 P}(\wv^T X - Y)(\wv^TX-Y)^T = \frac{1}{2} \left(\wv^T \Sigma^{xx} \wv - \wv^T\Sigma^{xy} - \Sigma^{xy,T} \wv + Y Y^T/P \right),
\label{eq_LR_loss}\eeq 
with the moment matrices
\beq \Sigma^{xx} =H = \frac{1}{P}X X^T,\qquad \Sigma^{xy} = \frac{1}{P} X Y^T.
\label{eq_cov_mats}\eeq 
Here, $H\in \reals^{N\times N}$ denotes the Hessian, which is identical to $\Sigma^{xx}$ for this model [see \cref{eq_LR_loss_atmin} below].
Over- and underparameterized regimes correspond to $N>P$ and $N<P$, respectively.
We will typically consider $\mu_x=0$; hence $\Sigma^{xx,xy}$ represent covariance matrices and, due to the i.i.d.\ assumption for $\xv^\mu$, in limit $P\to\infty$ one has $H_{ii} = \sigma_x^2$, while $H_{ij}=0$ for $i\neq j$.
In the following we make use of the Moore-Penrose pseudoinverse $\Acal^+$ of a matrix $\Acal$. The pseudoinverse can be expressed as $\Acal^+ = (\Acal^T \Acal)^{-1} \Acal^T$ if $\Acal$ has linearly independent columns, and as $\Acal^+ = \Acal^T (\Acal \Acal^T)^{-1}$ if $\Acal$ has linearly independent rows.

Using the elementary derivative rule $\pd(\xv^T \av)/\pd \xv=\av = \pd(\av^T \xv)/\pd\xv$, the condition for the minimum of the loss can be written as 
\beq \nabla_\wv L = \Sigma^{xx} \wv^* - \Sigma^{xy}=0.
\label{eq_LR_loss_grad}\eeq
In the underparameterized case, the solution readily follows as $\wv^*\st{up} = (\Sigma^{xx})^{-1}\Sigma^{yx}$. By contrast, in the overparameterized case, $\Sigma^{xx}$ is not full rank, and hence infinitely many solutions can be obtained by adding an arbitrary vector of the nullspace of $\Sigma^{xx}$ to any particular solution $w^*$. Taking advantage of this freedom, one thus typically considers the minimum norm solution obtained from the constrained minimization problem $\tilde L = ||\wv||^2 - (\wv^T X-Y)\boldsymbol\kappa$, where $\boldsymbol\kappa\in\reals^{P}$ is a vector of Lagrange multipliers \cite{schaeffer_double_2023}. The solution in the overparameterized case then follows as $\wv^*\st{op} = X (X^T X)^{-1} Y^T = (X X^T)^+ XY^T$, where we used the fact that the so-called gram matrix $X^T X\in \reals^{P\times P}$ now has full rank and applied the general identity $X(X^T X)^+ = (X X^T)^+ X$ of the pseudoinverse.
Defining 
\beq J = \frac{1}{P} \sum_\mu^P \xv^\mu \epsilon^\mu= \frac{1}{P} X \boldsymbol\varepsilon^T \in \reals^{N},
\label{eq_XErr}\eeq 
the minimum-norm solution of the linear regression problem can be compactly written as 
\beq \wv^* = (\Sigma^{xx})^+ \Sigma^{yx} = \begin{cases}
	(X X^T)^{-1} X Y^T = \uv + H^{-1} J, \qquad &N<P \quad \text{(underpar.)} \\
	X(X^T X)^{-1} Y = H^+ H \uv + H^+ J, \qquad &N>P. \quad \text{(overpar.)}
\end{cases}
\label{eq_LR_sol}\eeq 
The quantity $H^+H$ can be understood as a projector onto the orthogonal complement of the nullspace of $H$. 

Notably, by using only general properties of the pseudo-inverse [and not \cref{eq_LR_sol}], the loss in \cref{eq_LR_loss} can be rewritten as \cite{ziyin_strength_2021}
\beq L(\wv) = \frac{1}{2} (\wv-\uv-H^+J)^T H (\wv-\uv-H^+ J) - \frac{1}{2} J^T H^+ J + \frac{1}{2P}\boldsymbol\varepsilon \boldsymbol\varepsilon^T.
\label{eq_LR_loss_exp}\eeq 
Using the symmetry of $H$ and the elementary relation $H^+ H H^+=H^+$ of the pseudoinverse, one can readily show that \cref{eq_LR_loss_exp} is minimized by \cref{eq_LR_sol}. Accordingly, \cref{eq_LR_loss_exp} can be reformulated as
\beq L(\wv) = \frac{1}{2} (\wv-\wv^*)^T H (\wv-\wv^*) - \frac{1}{2} J^T H^+ J + \frac{1}{2P}\boldsymbol\varepsilon \boldsymbol\varepsilon^T,
\label{eq_LR_loss_atmin}
\eeq 
making the minimizing property of $\wv^*$ apparent.
In the overparameterized case, the last two terms in \cref{eq_LR_loss_atmin} cancel, such that $L(\wv^*)=0$, showing that the model is able to perfectly fit the data generated via \cref{eq_teacher_y}.
This cancellation can be proven by using the SVD $X=VD^{1/2}U^T$ ($V\in \reals^{N\times N}$, $D^{1/2}\in \reals^{N\times P}$, $U\in \reals^{P\times P}$, with $V$, $U$ being orthogonal and $D^{1/2}$ having $\min(P,N)$ singular values on the diagonal) and the relation $X^+ = U(D^{1/2})^+ V^T$, which implies that $X^T(X X^T)^+ X = \Imat_{P\times P}$. By contrast, the latter relation does not hold in the underparameterized case, since $U \Imat_{P\times P}^N U^T \neq \Imat_{P\times P}$ if $N<P$, where $\Imat_{P\times P}^N$ is the unit matrix with the last $P-N$ elements on the diagonal being zero.

\subsection{Dynamics}

Next, we consider the continuum approximation of the stochastic gradient descent, i.e., the \emph{stochastic gradient flow} (SGF), which we will use to study the spectrum of the noise and its dependence on the number of parameters and on the label noise. 
We consider the stationary state at the end of training, where the weights fluctuate around the solution $\wv^*$. 

\subsubsection{SGD and continuum approximation}

With the mini-batch loss ($S=|\Bcal|$ is the mini-batch size) 
\beq L_B(\wv) = \frac{1}{S}\sum_{\mu\in\Bcal} \ell^\mu(\wv), \qquad \ell^\mu(\wv)=\onehalf (\wv^T \xv^\mu-y^\mu)^2,
\label{eq_MB_loss}\eeq 
and the learning rate $\lambda$, the weights evolve according to SGD as 
\beq \wv(\tau+1) = \wv(\tau) - \lambda \nabla_w L_B(\wv(\tau)) = \wv(\tau) - \lambda \nabla L(\wv(\tau)) - \lambda\, \xib(\wv(\tau)),
\label{eq_SGD_disc}\eeq 
where $\tau$ denotes the discrete time step.
The mini-batch noise $\xib$ is defined as
\beq \boldsymbol\xi(\wv) \equiv \nabla L_B(\wv) - \nabla L(\wv)
\label{eq_MB_noise}\eeq 
and fulfils $\nabla L = \bra \nabla L_B\ket_\Bcal$, where $\bra\cdot\ket_B$ denotes the expectation over the full batch.

In order to proceed to the continuum limit, we follow \cite{kunin_limiting_2021, li_stochastic_2017, jastrzebski_three_2018, chaudhari_stochastic_2017} and assume the learning rate $\lambda$ to be sufficiently small, such that \cref{eq_SGD_disc} can be approximated, in the weak sense, by a SDE (Langevin equation): 
\beq \pd_t \wv(t) = -\nabla L(\wv(t)) +  \sqrt{\lambda} \,\xib(t),\qquad \nabla L(\wv)=H(\wv-\wv^*),\qquad \bra \boldsymbol\xi(t) \boldsymbol\xi(t')^T \ket_\Bcal = C(\wv) \delta(t-t')
\label{eq_SGD_Langevin}\eeq 
where $t=\tau/\lambda$ is a new time variable.
The noise $\xib(t)$ has weight-dependent Gaussian correlations in weight space, as characterized by the correlation matrix $C(\wv)$, which is further analyzed below. The noise is interpreted in an Ito sense, since, according to \cref{eq_SGD_disc}, $\wv(\tau)$ is independent of $\wv(\tau+1)$ \cite{gardiner_stochastic_2009}. Although the SGD-SDE mapping is not strictly rigorous \cite{yaida_fluctuationdissipation_2019, li_validity_2021}, we find that the results obtained from \cref{eq_SGD_Langevin,eq_SGD_disc} agree well even for rather large learning rates. 

\subsubsection{Noise covariance}
\label{sec_noise_covar}

According to \cref{eq_MB_noise}, the covariance matrix of the noise, averaged over batches (with replacement), is given by (see \cref{app_cov_MB})
\beq C = \frac{1}{S}\left[ \frac{1}{P} \sum_{\mu=1}^P (\nabla \ell^\mu) (\nabla \ell^\mu)^T - (\nabla L) (\nabla L)^T \right].
\label{eq_noise_cov_gen}\eeq 
The expression in the case of batching without replacement differs by the prefactor and vanishes for $S=P$, but approaches the above when $S\ll P$.
Note that for batching with replacement, which we use mainly in the present work, gradient noise is nonzero even if $S=P$.
In the case of linear regression, we have $\nabla_w\ell^\mu=(\wv^T \xv^\mu -y^\mu)\xv^\mu$ [see \cref{eq_MB_loss}]. 
Using \cref{eq_teacher_y}, the covariance takes the form:
\beq C_{ab}=\frac{1}{S}\left[ \frac{1}{P}\sum_\mu^P [\bar \wv^T\xv^\mu - \epsilon^\mu]^2 x^\mu_a x^\mu_b -H\bar \wv \bar \wv^T H \right] ,\qquad \bar \wv \equiv \wv-\uv.
\label{eq_LR_noise_cov_gen}\eeq 

This result shows that, beside the external source of randomness stemming from the mini-batching of input data $\xv^\mu$ and the label noise $\epsilon^\mu$, also the fluctuations of the weights $\wv$ contribute to the SGD-noise, rendering it state dependent \cite{meng_dynamic_2020a, liu_noise_2021}.
Estimating the typical magnitudes of the various terms in \cref{eq_LR_noise_cov_gen} (see \cref{app_addnoise}) indicates that the $w$-dependent terms dominate if $|| \bar \wv^2|| \sigma_x^2 \gg \sigma_\epsilon^2$. In the absence of label noise ($\sigma_\epsilon^2=0$), this condition is fulfilled at all times.
By contrast, if $\sigma_\epsilon^2> 0$, we write $\wv = \wv^* + \delta \wv$, where $\delta\wv$ denotes the weight fluctuations around the stationary solution $\wv^*$ [\cref{eq_LR_sol}]. 
Since the typical stationary variance $\bra\delta w_i^2\ket  \sim (\lambda/S)\sigma_\epsilon^2$ [cf.\ \cref{eq_wts_cov_conv} below and \cref{app_addnoise}] and $||H^{-1}J||^2\sim \sigma_\epsilon^2/\sigma_x^2$, multiplicative noise terms will dominate in the stationary state only if $(\lambda/S)\sigma_x^2 \gg 1$.
In the case of non-vanishing label noise ($\sigma_\epsilon^2>0$), the additive noise approximation is therefore warranted for typical choices of model parameters and normalized data.
Exceptions can arise at early stages of the training, where $||\bar \wv||$ can be large.
In the following, we will study in more detail the various approximations to the SGD noise.

\begin{figure}[t]\centering
    \subfigure[]{\includegraphics[width=0.32\linewidth]{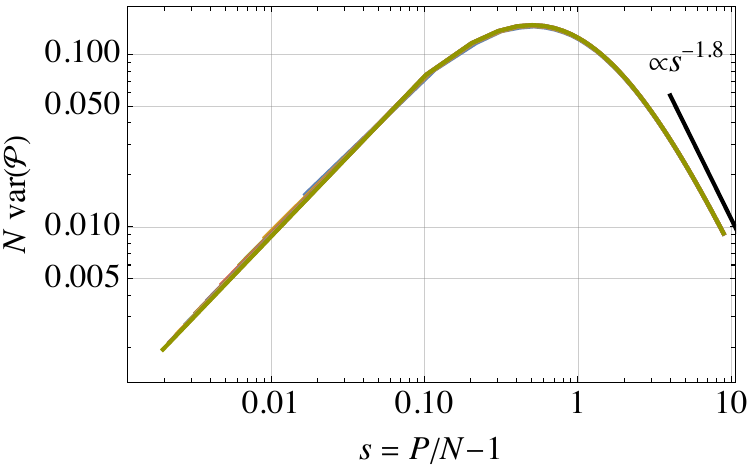} \label{fig_noise_proj}}\qquad  
    \subfigure[]{\includegraphics[width=0.32\linewidth]{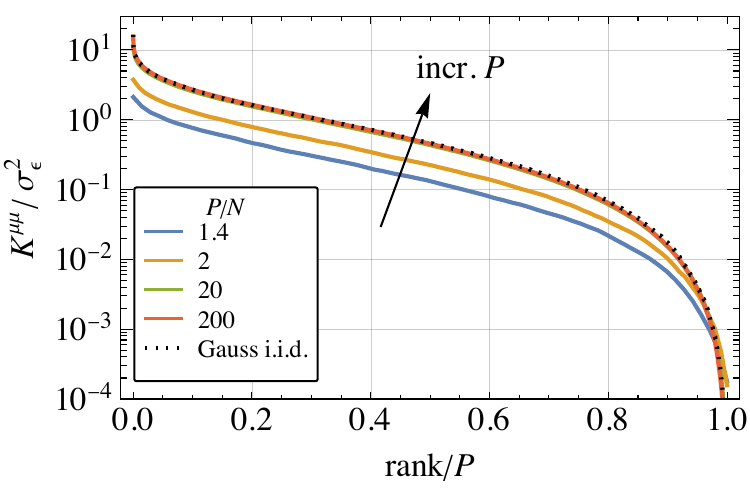}\label{fig_noise_corrmat}}
    \caption{Deviations from the Hessian approximation of the noise covariance of SGD of a linear network [\cref{eq_LR_noise_cov}]. (a) Variance of the matrix elements of $\Pcal=X^+X$, which enters in \cref{eq_noise_correct}, as a function of the sample number $P$ rescaled by the input dimension $N$. Curves for various $N$ and $P$ superimpose when expressed in appropriately scaled coordinates. (b) Entries, ordered by magnitude, of the diagonal random matrix $K$ [\cref{eq_noise_correct}], which determines the covariance of the SGD noise in \cref{eq_LR_noise_cov}. The various solid curves correspond to sample numbers ranging between $P=70$ and $P=10000$ (from bottom to top) with fixed input dimension $N=50$,  averaged over 100 realizations of $K$. The dotted black curve represents the approximation $K^{\mu\mu} = (\boldsymbol\epsilon^\mu)^2$, where $\boldsymbol\epsilon^\mu$ is a vector of independent Gaussian random numbers of zero mean and variance $\sigma_\epsilon^2$. The scaling of $\var(\Pcal)\propto N^{-1}s^{-1.8}$ observed in (a) ensures that the elements of the diagonal matrix $K$, which characterizes the deviation of the noise from the Hessian [see \cref{eq_LR_noise_cov}], are Gaussian i.i.d. for $s\to\infty$.}
\end{figure}

\textit{Additive noise approximation for finite sample number ($P<\infty$).} The additive noise approximation consists of evaluating $C$ [\cref{eq_LR_noise_cov_gen}] on the stationary solution $\wv= \wv^*$ [\cref{eq_LR_sol}], assuming low noise magnitude ($S\gg 1$).
For an overparameterized network, each individual loss gradient vanishes for $\wv=\wv^*$, i.e., $\nabla_\wv \ell^\mu(\wv^*) =0$ \footnote{This relation, which is exact in our setup, is also assumed to hold in general overparameterized deep networks \cite{ma_power_2018}.}. 
Consequently, in this case, the noise and the weight fluctuations vanish in the steady-state [see also the discussion after \cref{eq_LR_loss_atmin}]:
\beq C=0 \quad \text{if $N>P$ (overparameterized)}.
\label{eq_zero_noise_OP}\eeq 
By contrast, in the underparameterized regime, only the sum over the total batch is zero: $\sum_\mu^P \nabla_\wv \ell^\mu=0$.
We remark that the analysis here also applies for output dimensions larger than one, i.e., $\yv^\mu\in \reals^{d_o}$ with $d_o>1$, since the different output channels are not coupled by the loss. 

Since generally $J\neq 0$ [see \cref{eq_XErr}] for finite sample number, \cref{eq_LR_noise_cov_gen} takes the form
\beq C_{ab} = \frac{1}{S}\left[ \frac{1}{P}\sum_\mu^P (J^T H^+ \xv^\mu - \epsilon^\mu)^2 x^\mu_a x^\mu_b\right]
= \frac{1}{S}(X K X^T)_{ab},
\label{eq_LR_noise_cov}\eeq 
with a diagonal matrix in sample space:
\beq K^{\mu\nu}=\frac{1}{P} \left[\sum_{j,k}^N \sum_\gamma^P \frac{1}{P} \epsilon^\gamma x_k^\gamma H^+_{kj} x^\mu_j - \epsilon^\mu \right]^2 \delta^{\mu\nu} = \frac{1}{P} \left[\left( \frac{1}{P} \boldsymbol\varepsilon X^T H^+ X  - \boldsymbol\varepsilon\right)^\mu\right]^2 \delta^{\mu\nu}.
\label{eq_noise_correct}\eeq 
Note that these expressions are general and do not rely on specific assumptions for the statistics of the data or the label noise.
In the case of fully random labels $y^\mu$, they apply with the replacement $\epsilon^\mu \to y^\mu$.
Similar forms of $C$ in various limits have also been obtained in \cite{ali_implicit_2020, chen_statistical_2021}. 

In the \emph{over}parameterized case, one readily shows, using identities for the pseudoinverse, that $X^T(XX^T)^+X=\Imat$, hence $K=0$. In the \emph{under}parameterized case, instead, one has $X^T(XX^T)^+X=X^+X=:\mathcal{P}$, which is a projector on the row space of $X$ and is independent of the variance of $X$. In the large sample regime $P\gg N$, a numerical analysis [see \cref{fig_noise_proj}] reveals that the variance of the elements of this matrix behaves as $\var(\Pcal)\propto P^{-1.8} N^{-1}$ (the variance is slightly larger on the diagonal), while the mean value along the diagonal $\bra \Pcal_{ii}\ket\simeq N/P$. For the purpose of estimating the variance of $\varepsilon\Pcal$, we write $\Pcal \simeq (N/P) \Imat_P + P^{-0.9}N^{-0.5}\tilde \Pcal$, where $\tilde\Pcal$ is a random matrix with i.i.d.\ entries of zero mean and unit variance. Accordingly, we infer $\var(\boldsymbol\varepsilon X^+ X)_{ij}\sim P \sigma_\epsilon^2 \var(\Pcal) \to 0$ as $P\to\infty$, which implies $K^{\mu\nu} \simeq (\boldsymbol\varepsilon^\mu)^2 \delta^{\mu\nu}$, i.e., in the relevant oversampled regime and for $P\sigma_\epsilon^2\gg 1$, $K$ essentially becomes a diagonal matrix of squared Gaussian random numbers. This is illustrated in \cref{fig_noise_corrmat}, where the diagonal entries of $K$ (ordered by magnitude) are displayed for various sample sizes $P$.

\begin{figure}[t]\centering
    \subfigure[]{\includegraphics[width=0.335\linewidth]{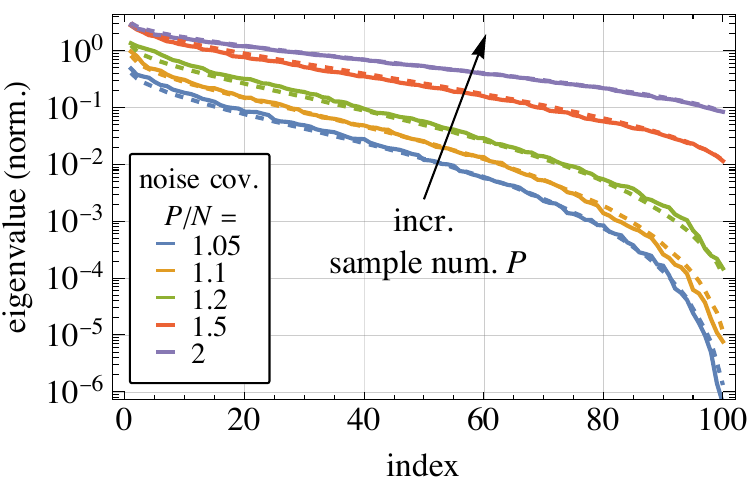} }\qquad 
    \subfigure[]{\includegraphics[width=0.34\linewidth]{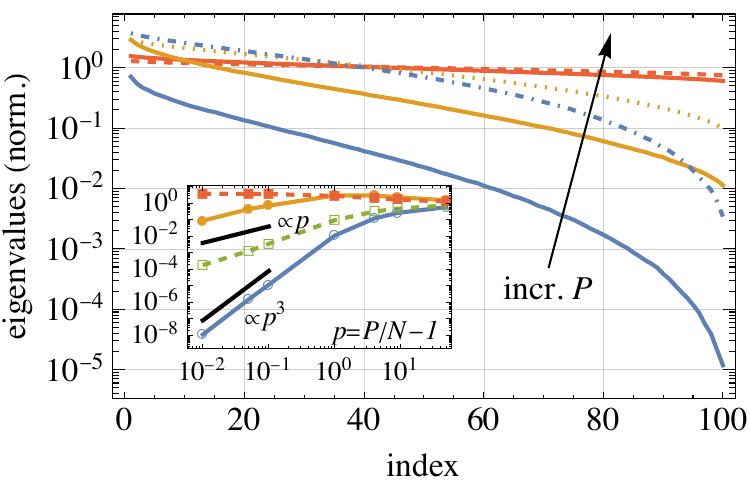} }\qquad 
    \subfigure[]{\includegraphics[width=0.22\linewidth]{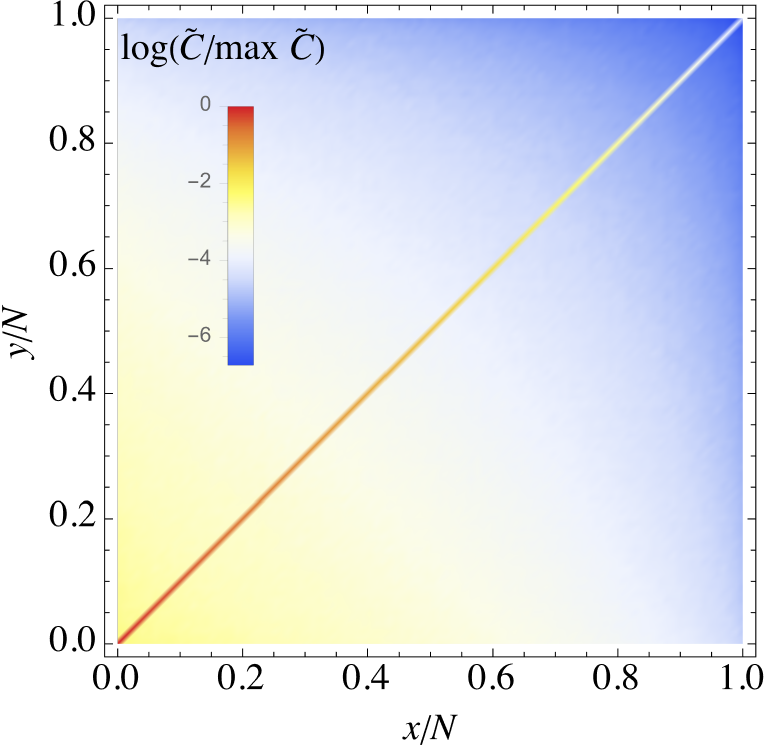} }
    \caption{Gradient noise of SGD for a linear network [see \cref{eq_LR_noise_cov_gen}]: anisotropy, dependence on sample number, and deviations from the Hessian approximation. (a) Eigenvalues of the gradient noise covariance matrix $C$, ordered by their magnitude, for varying number of samples $P$ (increasing from bottom to top) in the underparameterized regime and at input dimension $N=100$. Numerical results obtained from SGD (single data realization with learning rate $\lambda=0.1$, mini-batch size $S=10$, label noise $\sigma_\epsilon^2=10^{-2}$; solid lines) are compared to the theoretical predictions obtained from \cref{eq_LR_noise_cov} (averaged over multiple data realizations). (b) Validity of Hessian approximation of the noise for $P/N\gg 1$: Eigenvalues of $C$ given by \cref{eq_LR_noise_cov} (solid lines) compared to those of \cref{eq_noise_cov_Hess} (broken lines, essentially corresponding to the eigenvalues of the Hessian $H$), for $P/N=1.1,2,50$ (from bottom to top right) at $N=100$. The inset shows the maximum and minimum eigenvalues of $C$ from \cref{eq_LR_noise_cov} (solid connecting lines) and \cref{eq_noise_cov_Hess} (dashed connecting lines) for varying $P$. Eigenvalues in (a,b) are normalized by the factor $\sigma_x^2 \sigma_\epsilon^2/S$, where $\sigma_x^2=1/N$ is the variance of the samples [see also \cref{eq_noise_cov_Hess}]. (c) Visualization of the noise covariance (for $N/P\simeq 0.67$) in the basis of the Hessian $H$ [\cref{eq_cov_mats}], i.e., $\tilde C = \bra |V^T C V|\ket_{x,\epsilon}$, where the average is over several data realizations and the matrix $V$ consists of the eigenvectors of $H$ as columns. The color range is scaled logarithmically and normalized to the maximum entry of $\tilde C$. The magnitude of the off-diagonal entries decrease upon increasing $P$. 
    }
    \label{fig_noise_evs}
\end{figure}

In \cref{fig_noise_evs}(a), the spectrum of the noise covariance matrix obtained from numerical experiments via mini-batch sampling in the stationary state is shown and compared to the theoretical prediction of \cref{eq_LR_noise_cov} averaged over the distributions of the data $x$ and the label noise $\epsilon$ (which we take here to be Gaussian i.i.d., see \cref{sec_statics}).
The good agreement of the two spectra supports the validity of the additive noise approximation in this regime.
As shown in \cref{fig_noise_evs}(b), the noise spectrum deviates in general from the Hessian approximation of \cref{eq_noise_cov_Hess}, but approaches the latter in the strongly underparameterized limit $P/N\to\infty$. This is also confirmed by the calculations leading to \cref{eq_noise_cov_Hess} below.
The deviation is caused by the intermixing between the data eigenmodes due to the label noise [see \cref{eq_LR_noise_cov}], as illustrated in \cref{fig_noise_evs}(c).
This becomes also clear by noting that, from the SVD of the design matrix $X=V D^{1/2} U^T$ (with $V\in \reals^{N\times N}$, $U\in \reals^{P\times N}$, and a positive-semidefinite diagonal matrix $D^{1/2}\in \reals^{N\times N}$), the noise covariance in the Hessian eigenbasis takes the form $V^T C V = \frac{1}{S} D^{1/2} U^T K U D^{1/2}$, which differs from $D/S$ since $U^T K U\neq \Imat$.

\textit{Large sample number limit ($P\to\infty$).} In this regime, using the Gaussian i.i.d.\ assumption of the data and the label noise, the matrix $C$ can be evaluated as \cite{ziyin_strength_2021} \footnote{It is useful to note that $(\nabla L) (\nabla L)^T = H\bar \wv \bar\wv^T H$, which cancels one such term stemming from the first term in \cref{eq_noise_cov_gen}.} 
\beq C = \frac{1}{S}\left[ H \bar \wv \bar \wv^T H + \tr(H \bar \wv \bar \wv^T)H + \sigma_\epsilon^2 H \right],\qquad P\to\infty,
\label{eq_noise_cov_largeN}\eeq 
which explicitly reveals the multiplicative character of the gradient noise.
Using the fact that $J= 0$ [see \cref{eq_XErr}] for $P\to\infty$ (since $x^\mu$ and $\epsilon^\mu$ are uncorrelated), we have $\wv\simeq \wv^*=\uv$ in the stationary state [see \cref{eq_LR_sol}]. 
Setting thus $\bar\wv =0$ in \cref{eq_noise_cov_largeN} renders the additive noise approximation
\beq C \simeq \frac{\sigma_\epsilon^2}{S} H, \qquad P\to\infty.
\label{eq_noise_cov_Hess}\eeq 
This direct proportionality between the gradient covariance and the Hessian is often assumed in the literature \cite{mori_powerlaw_2022, kunin_limiting_2021, jastrzebski_three_2018}.

\textit{Multiplicative noise process for $\sigma_\epsilon^2=0$.}
Returning to the general case $\wv\neq \wv^*$, further insight into the nature of the stochastic process associated with SGD can be gained in the case of vanishing label noise $\varepsilon=0$. In this situation, \cref{eq_noise_cov_gen} becomes
\beq C_{ab} = \frac{1}{S}\left[ \frac{1}{P}\sum_\mu^P \bar w_i x_i^\mu x_a^\mu x_b^\mu x_j^\mu \bar w_j - \left(\frac{1}{P}\bar w_i \sum_\mu x_i^\mu x_a^\mu \right)\left(\frac{1}{P}\sum_\mu x_b^\mu x_j^\mu \bar w_j\right)\right] = \frac{1}{S} \cov_x [(\bar \wv\cdot \xv) \xv]_{ab}. 
\eeq 
Accordingly, \cref{eq_SGD_Langevin} can be formulated as 
\beq \pd_t\bar\wv = -H\bar \wv + \sqrt{\frac{\lambda}{S}} \bar \wv^T Z,
\label{eq_SGD_nolabeln}\eeq 
with a matrix-valued Gaussian white noise $Z_{ij}$ with covariance ($\Delta x_i x_j \equiv x_i x_j - \bra x_i x_j\ket$)
\beq  \bra Z_{ij}(t)Z_{kl}(t')\ket  = \bra (\Delta x_i x_j) (\Delta x_k x_l)\ket \delta(t-t')  = (\bra x_i x_k\ket\bra x_j x_l\ket + \bra x_i x_l\ket\bra x_j x_k\ket) \delta(t-t').
\label{eq_SGD_nolabeln_ncov}\eeq 
In the infinite sample limit ($P\to\infty$), the i.i.d.\ assumption (as well as $\mu_x=0$) for the $x_i$ implies $\bra Z_{ij}(t) Z_{kl}(t)\ket = (\delta_{ik}\delta_{jl} + \delta_{il}\delta_{jk}) \bra x^2\ket \delta(t-t')$ or, equivalently, $C_{ab} = S^{-1}(\bar w_a \bar w_b + \bar \wv^2 \delta_{ab}) \bra x^2\ket^2$.
The noise thus generally couples the different weights $\bar w_i$ during the evolution of \cref{eq_SGD_nolabeln}, even for whitened data (where $H=\Imat_N$). 
In the case of $N=1$ dimensions (but arbitrary finite $P$), the noise covariance simplifies to $\bra (\Delta x^2)^2\ket = \bra x^4\ket - \bra x^2\ket^2 = 2\bra x^2\ket^2$ and \cref{eq_SGD_nolabeln} reduces to the standard geometric Brownian motion process \cite{ali_implicit_2020}
\beq \pd_t \bar w = -\bra x^2\ket \bar w + \sqrt{\frac{2\lambda}{S}} \bra x^2\ket \bar w \zeta, \qquad \bra\zeta(t)\zeta(t') = \delta(t-t'),
\eeq 
where $\zeta$ is a Gaussian white noise.
The mean and variance of $\bar w$ for this process are given by $\bra \bar w(t)\ket = \bar w(0) e^{-\bra x^2\ket t}$ and $\bra \bar w(t)^2\ket = \bar w(0)^2 \left[ e^{2\bra x^2\ket \left(\frac{\lambda}{S}\bra x^2\ket -1 \right)t} - e^{-2\bra x^2\ket t} \right]$, showing that both converge to zero provided $\bra x^2\ket < S/\lambda$ (which is fulfilled for typical parameter choices).  
If $\bra x^2\ket=S/\lambda$ and $\bar w(0)^2=\lambda \bra x^2\ket/S=1$, mean and variance become identical to those of an Ornstein-Uhlenbeck process.

\subsubsection{Relaxational dynamics}

\begin{figure}[t]\centering
    {\includegraphics[width=0.34\linewidth]{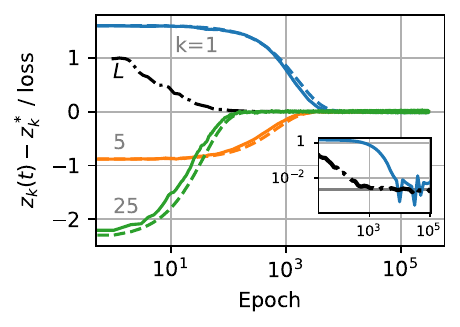} }
    \caption{Exemplary relaxation behavior of the loss $L(\zv)$ (dash-dotted curve) and the weight modes $z_k(t)=V^T_{kj} w_j$ [for $k=1,5,25$, see \cref{eq_SGD_Langevin_transf}] towards the solution $z_k^*$ (single run). Modes obtained from SGD experiments (solid lines) are compared to theoretical predictions $\bra z_k(t)\ket$ (\cref{eq_LR_modes_dyn}, averaged over the noise; dashed), for a slightly underparameterized linear net with $P/N\simeq 1.04$. The inset shows the loss and the mode $|z_0(t)|$ in double-logarithmic representation. The gray line in the inset represents the theoretical prediction $L^*$ for the final loss. Parameters used for SGD experiments: learning rate $\lambda=0.1$, label noise $\sigma_\epsilon^2=10^{-2}$, input dimension $N=50$, sample number $P=52$, batch size $S=10$. }   
    \label{fig_loss_relax}
\end{figure}

In order to study the dynamics induced by the SGD, we project the weights onto the eigenmodes of the data.
Assuming the SVD of the design matrix to be $X=P^{1/2} V D^{1/2} U^T$, with $V\in \reals^{N\times N}$, $U\in \reals^{P\times N}$, and a positive-semidefinite diagonal matrix $D^{1/2}\in \reals^{N\times N}$ (the sample number $P$ is introduced in order for the entries of $D$ to coincide with the eigenvalues of $H$), the Langevin \cref{eq_SGD_Langevin}, within the additive noise approximation, can be transformed to
\beq \dot \zv = -D(\zv-\zv^*) + \sqrt{\lambda}\boldsymbol\eta,\qquad \bra \boldsymbol\eta \boldsymbol\eta^T\ket = \frac{P}{S} D^{1/2} U^T K U D^{1/2},
\label{eq_SGD_Langevin_transf}\eeq 
where $\zv\equiv V^T \wv\in \reals^N$ is the new dynamical variable, $\zv^* \equiv V^T \wv^*$ are the transformed solution weights [see \cref{eq_LR_sol}], and $\boldsymbol\eta=V^T\boldsymbol\xi$ is a new noise (Gaussian distributed with zero mean). Analogously, the loss [\cref{eq_LR_loss_atmin}] can be expressed as 
\beq L = \onehalf (\zv-\zv^*)^T D (\zv-\zv^*) + L^* = \onehalf \sum_k^N (z_k-z_k^*)^2 \omega_k + L^*,
\label{eq_LR_loss_modes}\eeq
with $\omega_k \equiv D_{kk}$ and $L^* \equiv L(\zv^*) = - \frac{1}{2} J^T H^+ J + \frac{1}{2P}\boldsymbol\varepsilon \boldsymbol\varepsilon^T = \frac{1}{2 P} \boldsymbol\varepsilon (\Imat_P - X^T (XX^T)^{-1} X )\boldsymbol\varepsilon^T$ denoting the loss of the fully trained model.
\Cref{eq_SGD_Langevin_transf} describes a multivariate Ornstein-Uhlenbeck process driven by correlated Gaussian noise. Owing to the presence of label noise, encoded in the matrix $K$ [see \cref{eq_noise_correct}], the covariance of the noise $\boldsymbol{\eta}$ is not diagonal but instead induces correlations between the weight modes $z_i$.
The solution of \cref{eq_SGD_Langevin_transf} is given by
\beq z_k(t) = z_k^* + e^{-\omega_k (t-t_0)} [z_k(t_0)-z_k^*]  + \int_{t_0}^t \d s\, e^{\omega_k(s-t)}\eta_k(s) .
\label{eq_LR_modes_dyn}\eeq 
where, henceforth, we assume the relaxation rates to be ordered as $\omega_0\leq \omega_1\leq \ldots$.
For times larger than the smallest relaxation time ($t\gg 1/\omega_0$), the influence of the initial value term $\zv(t_0)-\zv^*$ has decayed and $\zv$ fluctuates around the equilibrium solution $\zv^*$.
Within the additive noise approximation considered here, the gradient descent (GD) solution follows by averaging $z_k(t)$ over the noise using $\bra \eta_k \ket=0$.
Near the interpolation threshold $N=P$, the eigenvalues of the Hessian $H$ become arbitrarily small, i.e., $\omega_0\to 0^+$, implying a diverging relation time \cite{advani_highdimensional_2020, rocks_memorizing_2020, singh_phenomenology_2022}. Specifically, for Gaussian data we consider here [see \cref{sec_statics}], the Marchenko-Pastur law of random matrix theory implies $\omega_0 = \sigma_x^2(\sqrt{P/N}-1)^2$ asymptotically for $P,N\to\infty$ with $P/N=\const$ \cite{bun_cleaning_2017,advani_highdimensional_2020} \footnote{In the overparameterized case ($P<N$), additionally a number of exactly zero eigenvalues appears.}.

The typical relaxation behavior of the modes $z_k(t)$ under SGD near the interpolation threshold is exemplified in \cref{fig_loss_relax}. We find that the relaxation behavior is overall well described by \cref{eq_LR_modes_dyn}, with minor deviations from the purely exponential relaxation for $P\simeq N$.
An interesting observation (see inset to \cref{fig_loss_relax}) is that the loss settles to the constant $L^*$ already for $t\ll \omega_0$, although the slowest modes $z_k$ have not yet all reached their final value.
This is a consequence of the presence of the factor $\omega_k$ in the loss [\cref{eq_LR_loss_modes}], due to which the contribution of modes for small $\omega_k$ can be hidden below the noise floor $L^*$, especially for large label noise $\sigma_\epsilon^2$.

\subsection{Steady-state properties}

In the following, we analyze the stationary covariance matrix of the weights induced by the SGF of a single-layer linear network. We focus, in particular, on the spectral properties of the weight fluctuations and their dependence on the sample number and the label noise. We then use these results to show that the IVFR is generally not fulfilled for a single-layer linear net. 

\subsubsection{Steady-state solution}

\label{sec_wts_fluct}

The Fokker-Planck equation for the distribution $p(t,\wv)$ induced by \cref{eq_SGD_Langevin} takes the form \cite{risken_fokkerplanck_1989,gardiner_stochastic_2009}:
\beq \pd_t p(t,\wv) = -\sum_i \frac{\pd}{\pd w_i} \Jcal_i,\qquad \Jcal_i = -\left[ \frac{\pd}{\pd w_i} L(\wv)\right] p(t,\wv) - \sum_j\frac{\pd}{\pd w_j} \left[\frac{\lambda }{2} C_{ij}(\wv) p(t,\wv)\right],
\label{eq_FP}\eeq 
with $L(\wv) = H(\wv-\wv^*)$.
We focus in the following on the $w$-independent noise defined by $C$ in \cref{eq_LR_noise_cov}, which, as discussed above, is expected to apply in the stationary state.
The steady-state covariance matrix of the weights $M \equiv \bra \delta \wv \delta \wv^T\ket = M^T$ (with $\delta\wv\equiv \wv-\wv^*$) is determined by the Lyapunov equation \cite{risken_fokkerplanck_1989, laub_matrix_2005, mandt_stochastic_2017, kunin_limiting_2021} 
\beq H M+ M H^T = \lambda C.
\label{eq_LR_Lyapunov}\eeq 
The most general solution of this equation fulfills $\nabla\cdot \Jcalv=0$, but not necessarily $\Jcalv= 0$, i.e., while the probability density remains stationary [see \cref{eq_FP}], the current $\Jcalv$ can still be non-vanishing (broken detailed balance). 
This situation is indeed realized for SGD and one obtains \cite{chaudhari_stochastic_2017, kunin_limiting_2021}:
\begin{align}
	M &\equiv \bra \delta \wv \delta \wv^T\ket = \frac{\lambda}{2} H^{-1} (C+Q), \label{eq_wts_cov} \\
	Q &= V \tilde Q V^T,\qquad  \tilde Q_{ij} = \frac{\omega_i - \omega_j}{\omega_i+\omega_j} (V^T C V)_{ij},\quad \text{where}\quad V \diag(\omega_i) V^T = H, \label{eq_wts_cov_Q}
\end{align}
i.e., $\omega_i$ are the eigenvalues of $H$ and the columns of $V$ contain the eigenvectors.
The stationary probability distribution and current follow as
\beq p\st{ss}(\wv) = \frac{1}{Z}\exp\left[ - \frac{1}{2}(\wv-\wv^*)^T M^{-1} (\wv-\wv^*)\right],\qquad \Jcalv\st{ss} = -QM^{-1}(\wv-\wv^*)p\st{ss}(\wv),
\label{eq_wts_pdf}\eeq 
with a normalization factor $Z$.
The inverse covariance matrix $M^{-1}$ can be interpreted as an effective potential $U(\wv)\equiv (1/2) \delta\wv^T M^{-1} \delta\wv$, which, due to the presence of the kinetic term $(C+Q)^{-1}$, differs in general from the loss \cite{chaudhari_stochastic_2017, xiong_stochastic_2022}.
The matrix $Q$ is anti-symmetric ($Q=-Q^T$) and determines a stationary current $\Jcalv\st{ss}$, which flows in planes of constant force, i.e., $\Jcalv\st{ss}\cdot (\nabla U)=0$ \cite{kwon_structure_2005, kunin_limiting_2021}.
In order to aid comparison between various conventions used, e.g., in numerical codes, we remark that, if the loss [\cref{eq_MB_loss}] is scaled by a factor $\alpha$, i.e., $\ell^\mu \to \alpha\ell^\mu$, the noise covariance [\cref{eq_noise_cov_gen}] scales as $C\to \alpha^2 C$  (analogously $Q\to \alpha^2 Q$) and the Hessian $H\to \alpha H$, which implies $M\to \alpha M$.

A solution fulfilling detailed balance can be obtained from straightforward integration of the condition $\Jcalv=0$ in \cref{eq_FP}, which yields $\onehalf \lambda C_{ij}\cdot \sfrac{\pd \ln p}{\pd w_j}  = -\sfrac{\pd L(\wv)}{\pd w_i}$ and eventually
\beq M=\bra \delta\wv \delta \wv^T\ket = \frac{\lambda}{2} H^{-1}C,\qquad Q=0. \qquad\text{(detailed balance)}
\label{eq_wts_cov_detbal}\eeq 
While, generally, detailed balance is broken for SGD, we find that \cref{eq_wts_cov_detbal} constitutes a reasonable approximation sufficiently deep in the underparameterized regime (specifically, for $P/N\gtrsim 2$, see discussion of \cref{fig_LR_weight_fluct} below). 
In fact, asymptotically for $P/N\to \infty$, one has $C\propto H$ [see \cref{eq_noise_cov_Hess}], which implies $V^T C V\propto \delta_{ij}$ and thus $Q=0$ in \cref{eq_wts_cov}.
In this limit detailed balance is thus exactly fulfilled and one obtains isotropic weight fluctuations \cite{kunin_limiting_2021, ziyin_strength_2021}, 
\beq M  = \frac{\lambda \sigma_\epsilon^2}{2S} \Imat,\qquad P/N\to \infty.
\label{eq_wts_cov_conv}\eeq

\begin{figure}[t]\centering
    \subfigure[]{\includegraphics[width=0.33\linewidth]{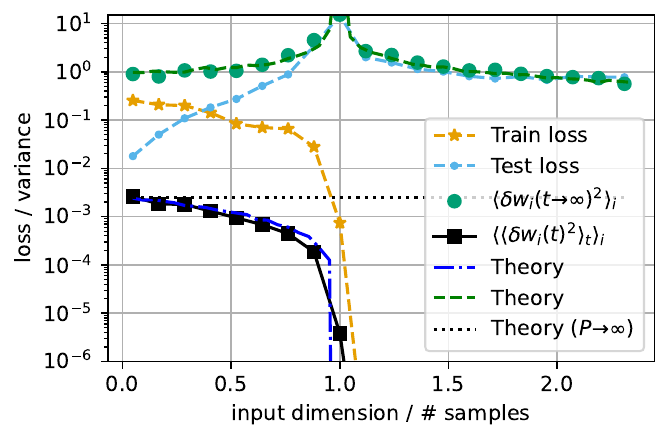} }
    \subfigure[]{\includegraphics[width=0.32\linewidth]{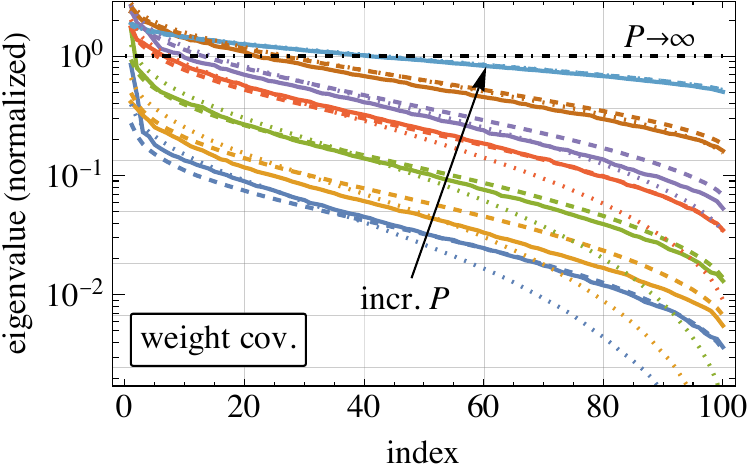} }
    \subfigure[]{\includegraphics[width=0.32\linewidth]{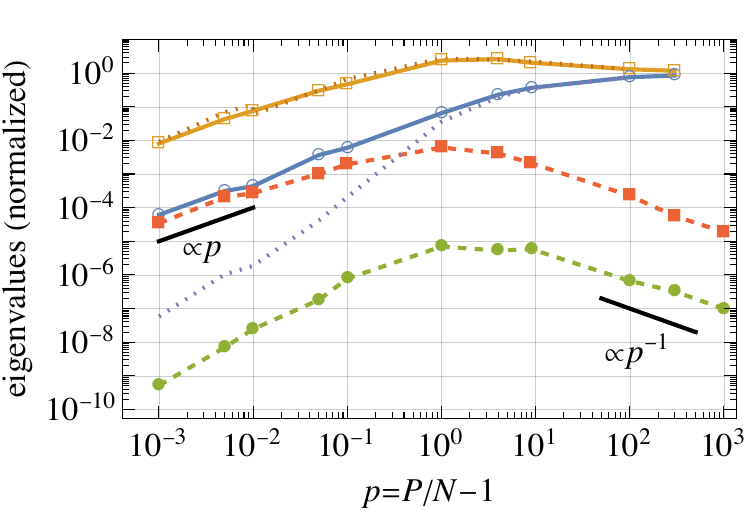} }
    \caption{Statistics of weight fluctuations for a linear network in the stationary state. (a) Behavior of the weight variance as a function of input dimension for $P=200$ samples. Simulation results (connected symbols) obtained for a single run of SGD with learning rate $\lambda=0.1$, label noise $\sigma_\epsilon^2=0.25$, and mini-batch size $S=10$ are compared to the theoretical predictions of \cref{eq_wts_cov_conv,eq_wts_cov} for $\bra\bra\delta w_i(t)^2\ket_t\ket_i = N^{-1}\sum_i M_{ii}$ (dotted and dash-dotted lines), which represents the temporal variance of the weights in the stationary state [where $\wv(t\to\infty)=\wv^*$, see \cref{eq_LR_sol}] averaged over all weights. The quantity $\bra\delta w_i(t\to\infty)^2\ket_i$ (filled circles and dashed line) represents the variance across all weights at an arbitrary but fixed time in the stationary state. The train and test losses are also shown for comparison. (b) Eigenvalues of the covariance matrix of the weights $M=\bra\delta\wv \delta\wv^T\ket$, ordered by their magnitude and normalized by $\lambda\sigma_\epsilon^2/(2S)$ [see \cref{eq_wts_cov_conv}], for varying number of samples ($P/N = 1.05, 1.1, 1.2, 1.5, 2.0, 4.0,10$, from bottom to top) in the underparameterized regime (for $\lambda=0.1$, $S=10$, and input dimension $N=100$). Numerical results obtained from SGD (solid lines) are compared to the theoretical predictions obtained from \cref{eq_wts_cov} (dashed lines; averaged over the multiple data realizations). The straight (dashed-dotted) black line represents the prediction of \cref{eq_wts_cov_conv}, while the dotted lines represent the theoretical eigenvalues under the detailed balance assumption [\cref{eq_wts_cov_detbal}]. (c) Smallest and largest eigenvalues of $M$ as given by the exact result \cref{eq_wts_cov} (solid lines) and the detailed balance approximation  \cref{eq_wts_cov_detbal} (dotted lines), normalized by $\lambda\sigma_\epsilon^2/(2S)$ and for varying sample numbers $P$. The dashed lines represent the smallest and largest absolute values of the eigenvalues of $Q$ [\cref{eq_wts_cov}] in the same normalization. We note that the impact of Q diminishes in the oversampled regime.}
    \label{fig_LR_weight_fluct}
\end{figure}

In \cref{fig_LR_weight_fluct}, the behavior of the weight fluctuations of a fully trained linear network is illustrated. As shown in panel (a), the steady-state variance $\bra \bra\delta w_i(t)^2\ket_t\ket_i$ obtained from SGD experiments (by computing the temporal variance of each weight $w_i$ and the averaging over all weights $i=1,\ldots,N$) agrees with the theoretical prediction given by the diagonal entries of $M$ in \cref{eq_wts_cov}. Notably, the variance is non-vanishing only for $N<P$, since an overparameterized network can fit the data in \cref{eq_teacher_y} exactly [see also the comment after \cref{eq_LR_noise_cov_gen}]. In the limit $N/P\to 0$, the variance approaches the isotropic result in \cref{eq_wts_cov_conv} (dotted horizontal line). For comparison, the train and test loss are included in the plot, where the latter shows a ``double descent'' peak at the interpolation threshold ($P=N$), characteristic for learning data with noisy labels  \cite{opper_ability_1990, krogh_generalization_1992, belkin_reconciling_2019, advani_highdimensional_2020, hastie_surprises_2022, nakkiran_deep_2019}. 
Beside the temporal variance, we show also the spatial variance of the weights $\bra\delta w_i(t\to\infty)^2\ket_i = \bra \delta (w_i^*)^2\ket$, which peaks at $P=N$ \cite{chen_statistical_2021}.
We recall that a stationary state is only reached for non-vanishing label noise. By contrast, for $\sigma_\epsilon=0$, the network can fit the teacher weights exactly such that the loss exponentially approaches zero (up to machine precision).

\Cref{fig_LR_weight_fluct}(b) illustrates eigenvalues of the weight covariance matrix $M=\bra\delta\wv \delta \wv^T\ket$, ordered by their magnitude, for various sample numbers $P$ in the underparameterized regime. The numerical results from SGD are well captured by \cref{eq_wts_cov} (dashed lines), while significant deviations from \cref{eq_wts_cov_conv} are noticeable, indicating broken detailed balance, except for large $P$.

In \cref{fig_LR_weight_fluct}(c), the smallest and largest eigenvalues of the weight covariance matrix $M$ [\cref{eq_wts_cov,eq_wts_cov_detbal}, solid and dotted lines, respectively] are shown as functions of the reduced sample number $P/N-1$. We infer that the detailed balance approximation for $M$ given in \cref{eq_wts_cov_detbal} is applicable for $P\gtrsim 2N$. 
This is also corroborated by the behavior of the eigenvalues of $Q$ [\cref{eq_wts_cov_Q}, dashed lines], which peak around $P\approx N$, and decay for larger sample numbers. (Note that the eigenvalues of the antisymmetric matrix $Q$ are purely imaginary and hence we plot their absolute value.)
From the gap between the minimum and maximum eigenvalues we finally conclude that the weight fluctuations become essentially isotropic [see \cref{eq_wts_cov_conv}] for $P\gtrsim 100N$, consistent with the behavior of the noise eigenvalues in \cref{fig_noise_evs}(b).

Before proceeding, we summarize some important steady-state properties for linear regression in the following \cref{tab_linreg_prop}. 

\begin{table}[h]
\centering
\begin{tabular}{|c|p{4.4cm}|p{5cm}|p{5cm}|}
\hline
  & student weights [\cref{eq_LR_sol}] & final loss [\cref{eq_LR_loss_atmin}] & temporal weight fluctuations  \\ \hline
underparameterized & approach teacher weights as $\epsilon\to 0$ & nonzero in general; vanishes $\propto \sigma_\epsilon^2$  & nonzero in general; vanish $\propto \sigma_\epsilon^2$ [\cref{eq_wts_cov_conv}] \\ \hline
overparameterized & differ from teacher weights & vanishes (also if $\sigma_\epsilon^2>0$) & vanish [\cref{eq_zero_noise_OP,eq_noise_cov_Hess}] \\ \hline
\end{tabular}
\caption{Properties of a linear network in the stationary (late-time) regime, trained on a teacher-student task [see \cref{eq_teacher_y}]. Note that these properties are independent of the sampling method (i.e., sampling with or without replacement). The variance of the target (label) noise is denoted by $\sigma_\epsilon^2$.}
\label{tab_linreg_prop}
\end{table}

\subsubsection{Loss perturbation}

\begin{figure}[b]\centering
    \subfigure[]{\includegraphics[width=0.3\linewidth]{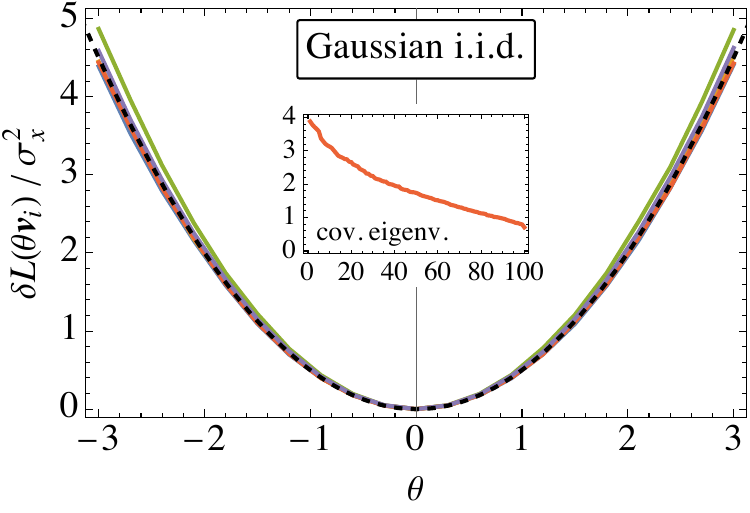} }\qquad 
    \subfigure[]{\includegraphics[width=0.31\linewidth]{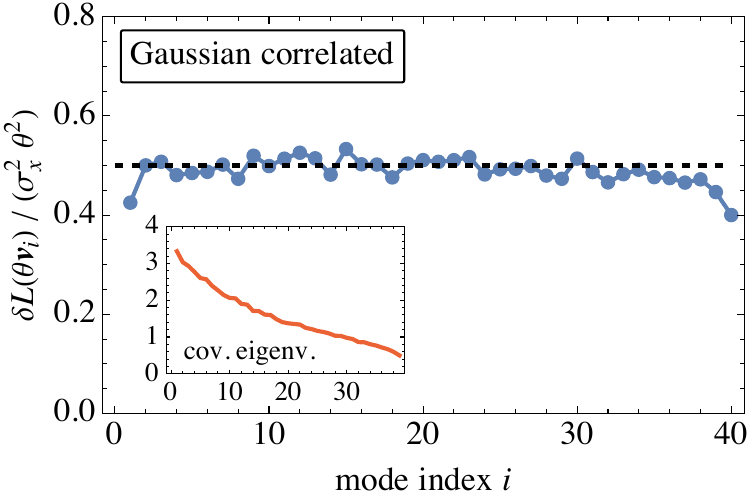} }
    \caption{Perturbing the linear regression loss [\cref{eq_LR_loss_atmin}] along eigenvectors of the empirical weight covariance matrix $M=\bra\delta\wv \delta \wv^T\ket$ obtained from SGD (learning rate $\lambda=0.1$, mini-batch size $S=10$) of a linear network. The insets show the eigenvalues of the weight covariance matrix (ordered by their magnitude vs. mode index $i$). (a) Perturbed loss curves $\delta L(\theta\vv_i)$ for various eigenvectors $\vv_i$ for  $N=100$ dimensional input and $P=2000$ Gaussian i.i.d.\ samples (single run). The simulation results superimpose on the theoretical prediction $\delta L = \onehalf \theta^2 \sigma_x^2$ (dashed curve) for large sample numbers. (b) Perturbed loss $\delta L(\theta \vv_i)/\theta^2$ at fixed (arbitrary) $\theta$ as a function of eigenmode index $i$ for $P=200$ correlated samples in $N=40$ dimensions obtained from a multivariate Gaussian defined by a random covariance matrix $\Sigma$ of Wishart type (i.e., $\Sigma\simeq MM^T$, with $M_{ij}\sim \Ncal(0,1)$). The black dashed line represents the theoretical prediction for i.i.d.\ data, while the connected symbols are obtained by averaging over several experimental runs.}
    \label{fig_loss_pert_LR}
\end{figure}

We now investigate whether the IVFR \cite{feng_inverse_2021} applies in the context of a single-layer linear network.
Perturbing the linear regession loss in \cref{eq_LR_loss_atmin} around the solution $\wv^*$ [\cref{eq_LR_sol}] renders
\beq \delta L(\theta \vv) = L(\wv^*+\theta \vv)-L(\wv^*) = \frac{\theta^2}{2} \vv^T H \vv ,
\label{eq_loss_pert_def}\eeq 
where $\vv$ is assumed to be an eigenvector of the weight covariance matrix $M$ [\cref{eq_wts_cov}]. Owing to the different structural forms of $M$ and $H$, the eigenvectors of these matrices can be considered as essentially unrelated here and we thus do not expect $\delta L(\theta\vv)$ to depend in any significant way on the covariance eigenvalues, even if the latter strongly vary in magnitude. This is confirmed by SGD simulations for Gaussian independent and Gaussian correlated data (see \cref{fig_loss_pert_LR}), as well as for purely random labels $y^\mu$. 

Analytic insight can be obtained in the large (infinite) sample number regime, where \cref{eq_wts_cov_conv} implies the eigenvectors $\vv\in \{\ev_i\}_{i=1,\ldots,N}$ with a common eigenvalue $\gamma=\lambda\sigma^2/S$. This yields $\delta L(\theta\ev_i) = \onehalf \theta^2 H_{ii} = \onehalf \theta^2 \bra x_i^2\ket$, i.e., the loss perturbation reveals the variance of the $i$th feature. Accordingly, unless one engineers a particular variance dependence on the feature index, the loss perturbation behaves in an essentially random and isotropic way. We thus conclude that the IVFR is not fulfilled for a typical single-layer linear network.

\section{Two-layer linear network}
\label{sec_2LNet}

We now transfer the Langevin formalism presented in the preceding section to a two-layer linear network, which will enable us to analyze the fluctuations of the weights in the individual layers. Since the dynamics of the weight matrices is now nonlinear, we develop a linearization around the deterministic solution. We will finally use these results in order to derive the IVFR. 
In order to facilitate analytical calculations, we will focus, where necessary, on the regime of large batch sizes, small learning rate, and large sample numbers.

\subsection{Model}
\begin{figure}[t]\centering
    {\includegraphics[height=0.21\linewidth]{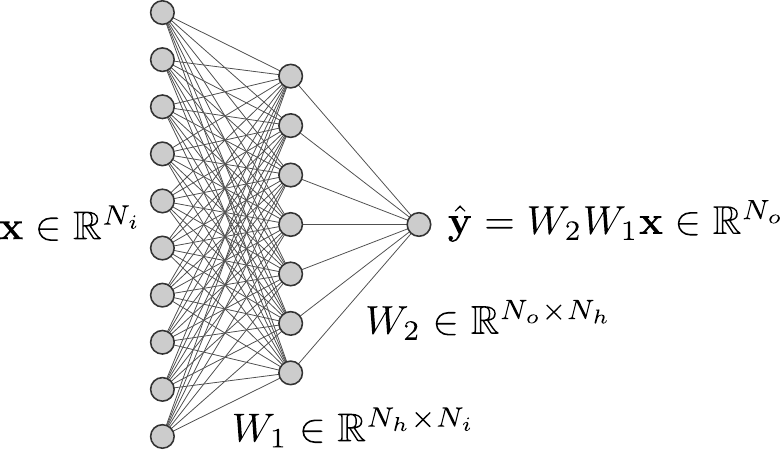} }
    \caption{Two-layer linear neural network model with input $\xv$, output $\hat\yv$, and weight matrices $W_1$, $W_2$. The targets are given as i.i.d.\ noise $\yv$ (see text).} 
    \label{fig_2lnet}
\end{figure}

We consider training data consisting of $P$ samples $\xv^\mu\in\reals^{N_i}$, $\yv^\mu\in\reals^{N_o}$ ($\mu=1,\ldots,P$), with i.i.d.\ Gaussian $x_i^\mu \sim \Ncal(0, \sigma_x^2)$ and $y_i^\mu\sim \Ncal(0,\sigma_y^2)$, which, for simplicity, have zero mean. 
Within the teacher-student framework used in \cref{eq_teacher_y}, the targets (``labels'') $\yv^\mu$ used here can be interpreted as pure noise, while the teacher weights identically vanish. This setup (see \cref{fig_2lnet}) is sufficient to investigate the stationary fluctuating weight dynamics, which is driven by label noise [cf.\ \cref{eq_LR_noise_cov}].
The data is conveniently organized in terms of the design matrices $X\in \reals^{N_i\times P}$, $Y\in\reals^{N_o\times P}$.
The (full-batch) loss of a linear 2-layer network is given by
\begin{equation}\begin{split}
L(W_1,W_2) &= \frac{1}{2 P} \sum_{\mu=1}^P\| \yv^\mu - W_2 W_1 \xv^\mu \|^2 = \frac{1}{2 P} \tr \left[(Y-w X)(Y-w X)^T\right] \\
&= \frac{1}{2} \tr \left( \Sigma_0^{xx} w^T w - w^T \Sigma_0^{yx} - \Sigma_0^{yx,T}w + YY^T/P \right), 
\label{eq_2L_loss}
\end{split}\end{equation}
where $W_1\in \reals^{N_h\times N_i}$, $W_2\in \reals^{N_o\times N_h}$ are the weight matrices, $w\equiv W_2 W_1\in \reals^{N_o\times N_i}$ denotes the product matrix, and
\beq \Sigma_0^{xx} = \frac{1}{P} \Xm \Xm^T ,\qquad \Sigma_0^{yx} = \frac{1}{P} \Ym \Xm^T 
\eeq 
are the full-batch input-input and output-input covariance matrices. 
The minimum of the loss is determined by the linear regression solution
\beq w^* \equiv (W_2 W_1)^* = \Sigma_0^{yx} (\Sigma_0^{xx})^{+} ,
\label{eq_2L_LRsol}\eeq 
which effectively fixes only $N\eff=N_o N_i$ degrees of freedom out of the total number of parameters $N\tot=N_h(N_i+N_o)$.
As was the case for a single-layer net [see \cref{eq_zero_noise_OP}], a deep linear network of the form considered here has non-vanishing weight fluctuations in the stationary state provided $N_i<P$, irrespective of the output dimension $N_o$.
We will thus restrict ourselves to this regime in the following, in which case the pseudo-inverse in \cref{eq_2L_LRsol} accordingly becomes the usual inverse.
In certain cases, where indicated, we will also assume $N_i\geq N_h\geq N_o$.

Introducing the mini-batch covariance matrices as
\beq \Sigma^{xx} = \frac{1}{S} \sum_{\mu\in\Bcal} \xv^{\mu} \xv^{\mu,T},\qquad \Sigma^{yx} = \frac{1}{S} \sum_{\mu\in\Bcal} \yv^{\mu} \xv^{\mu,T},
\eeq 
where $\Bcal$ denotes the set of sample indices in the current mini-batch (of size $S=|\Bcal|$), the SGD equations take the form 
\begin{subequations}
\begin{align} 
\frac{1}{\lambda}\left[ W_1(\tau+1)-W_1(\tau)\right] &= - \frac{\pd L(w,\tau)}{\pd W_1} =  W_2^T\left(\Sigma^{yx} - W_2 W_1 \Sigma^{xx}\right), \\
\frac{1}{\lambda}\left[ W_2(\tau+1)-W_2(\tau)\right] &= - \frac{\pd L(w,\tau)}{\pd W_2} =  \left(\Sigma^{yx} - W_2 W_1 \Sigma^{xx}\right) W^T_1 ,
\end{align}\label{eq_2L_SGD}
\end{subequations}
with the learning rate $\lambda$.

The full-batch dynamics of \cref{eq_2L_SGD} in the limit of small $\lambda$ (\emph{gradient flow}) has been analyzed previously \cite{saxe_exact_2013,saxe_mathematical_2019,braun_exact_2022,gunasekar_implicit_2017}. 
We will derive the continuum limit of \cref{eq_2L_SGD} in the next section, but already note here that, in the case of gradient flow with an explicitly time-dependent loss $L$, one can show that the following quantity is conserved for a linear network ($t=\tau/\lambda$): 
\beq W_2^T(t) W_2(t) - W_1(t) W_1^T(t) = \const.
\label{eq_2L_wtcons}\eeq 
The proof of this relation follows the same steps as in \cite{arora_optimization_2018}, where it was demonstrated for linear networks with losses that do not explicitly depend on time. 
When the weights are initialized such that the constant matrix on the r.h.s.\ of \cref{eq_2L_wtcons} vanishes, the weights are conventionally called \emph{zero-balanced}.

Notably, for whitened inputs $\Sigma_0^{xx}\propto \Imat$ and zero-balanced weights, an explicit solution of this gradient flow can be constructed \cite{braun_exact_2022} \footnote{The proof presented in Appendix C of \cite{braun_exact_2022} should apply for arbitrary $N_i$ and $N_o$.}:
Denoting the compact singular value decomposition of $\Sigma_0^{yx}$ by 
\beq \Sigma_0^{yx} = \tilde U \tilde S \tilde V^T, \qquad \text{with}\qquad \tilde U\in\reals^{N_o\times N_m}, \tilde S\in\reals^{N_m\times N_m}, \tilde V\in\reals^{N_i\times N_m},
\eeq 
where $\tilde S$ is a diagonal matrix with $N_m$ nonzero eigenvalues on its diagonal, the late-time solution can be expressed as
\beq W_1^{0,T} W_1^0 = \tilde V \tilde S \tilde V^T,\qquad W_2^0 W_2^{0,T} = \tilde U \tilde S \tilde U^T. \qquad \text{(zero-balanced)}
\label{eq_2L_wtsol_zbal}\eeq 
However, explicit expressions for the individual full-batch $W_i^0$ are not available. Unless otherwise mentioned, we will in the following not assume any special initial conditions.

\subsection{Continuum description of SGD}

We now derive the continuum limit of the SGD dynamics in each layer, which renders two coupled nonlinear SDEs. A crucial step is to determine the covariances of the effective noise terms induced by the mini-batching. In order to analyze the nonlinear SDEs in the stationary state, we focus on the large batch-regime, where fluctuations around the deterministic gradient flow solution are small.

\subsubsection{Noise sources and stochastic gradient flow}

Since the only source of randomness enters via the mini-batching of $\Sigma^{xx,yx}$ \cite{han_fluctuationdissipationtype_2021}, we have
\beq \Sigma^{xx} = \Sigma^{xx}_0 + \delta \Sigma^{xx}, \qquad \Sigma^{yx} = \Sigma^{yx}_0 + \delta \Sigma^{yx},
\eeq 
where we note that $\Sigma_0^{xx} = \bra \Sigma^{xx}\ket_\Bcal$ and $\Sigma_0^{yx} = \bra \Sigma^{yx}\ket_\Bcal$.
Accordingly, the noise terms are defined as
\beq \delta \Sigma^{xx} = \Sigma^{xx} - \Sigma_0^{xx},\qquad \delta \Sigma^{yx} = \Sigma^{yx} - \Sigma_0^{yx},
\eeq 
and their covariances can be expressed (using \cref{app_cov_MB}) by a four-dimensional tensor:
\beq\begin{split} \mathrm{cov}(\delta \Sigma^{xx})_{ij,kl} \equiv \bra \delta\Sigma^{xx}_{ij} \delta\Sigma^{xx}_{kl}\ket_\Bcal &=  \left\bra \left( \frac{1}{S} \sum_{\mu} x^\mu_i x^\mu_j -\Sigma_{0,ij}^{xx}\right) \left(\frac{1}{S} \sum_{\nu} x^\nu_k x^\nu_l - \Sigma_{0,kl}^{xx}\right) \right\ket_\Bcal  \\
&= \frac{1}{S}\left[ \frac{1}{P} \sum_{\mu=1}^P x^\mu_i x^\mu_j x^\mu_k x^\mu_l - \Sigma_{0,ij}^{xx} \Sigma_{0,kl}^{xx} \right].
\end{split}\label{eq_2L_cov_sigmaXX}\eeq 
In general, a four-dimensional tensor $\Mcal_{ijkl}$ can be reshaped into a conventional two-dimensional matrix $\mcal$ by defining $\mcal_{ab}=[\Mcal_{ij,kl}]_{a=(j-1)N_i+i,b=(l-1)N_k+k}$ where $N_i$ and $N_k$ are the sizes of dimensions indexed by $i$ and $k$, respectively. This definition is consistent with the usual vectorization $\mcal=\vect(\Acal)\vect(\Bcal)^T$ in the case where $\Mcal_{ijkl}=\Acal_{ij}\Bcal_{kl}$, which applies to the covariance tensor.
The vectorization operation $\vect(\Acal)$ is defined by stacking the columns of the matrix $\Acal$ on top of each other to form a column vector.

In order to arrive at tractable expressions, we evaluate the noise covariance in the large sample number limit ($P\to\infty$), which is analogous to the Hessian approximation given by \cref{eq_noise_cov_Hess}. Practically, the Hessian approximation for the noise applies already for $P\gtrsim 100N$, as discussed in \cref{sec_wts_fluct}.
In this limit, the first term in the square brackets in \cref{eq_2L_cov_sigmaXX} converges to a batch average, such that Wick's rule for products of Gaussian variables can be applied:
$\frac{1}{P} \sum_{\mu=1}^P x^\mu_i x^\mu_j x^\mu_k x^\mu_l \simeq \bra x_i x_j x_k x_l\ket_\Bcal = \bra x_i x_j\ket_\Bcal \bra x_k x_l\ket_\Bcal + \bra x_i x_k\ket_\Bcal \bra x_j x_l\ket_\Bcal + \bra x_i x_l\ket_\Bcal \bra x_j x_k\ket_\Bcal$. Hence, altogether one obtains
\begin{subequations}
\begin{align}
\mathrm{cov}(\delta\Sigma^{xx})_{ij,kl} &= \frac{1}{S} \left[ \Sigma_{0,ik}^{xx}\Sigma_{0,jl}^{xx} + \Sigma_{0,il}^{xx}\Sigma_{0,jk}^{xx}\right], \\
\mathrm{cov}(\delta\Sigma^{yx})_{ij,kl} &= \frac{1}{S} \left[ \bra y_i^2\ket \delta_{ik} \Sigma_{0,jl}^{xx} + \Sigma_{0,il}^{yx}\Sigma_{0,kj}^{yx}\right], \\
\mathrm{cov}(\delta\Sigma^{xx}_{ij} ,\delta\Sigma^{yx}_{kl}) &= \frac{1}{S} \left[ \Sigma^{yx}_{0,ki} \Sigma_{0,jl}^{xx} + \Sigma_{0,il}^{xx}\Sigma_{0,kj}^{yx}\right], 
\end{align}	\label{eq_2L_cov_largeP}
\end{subequations}
\hspace{-0.24cm} where we used the factor that the $y_i$ are uncorrelated zero mean random variables.
In the case of whitened inputs, i.e.\ 
\beq \Sigma_0^{xx}=\sigma_x^2 \Imat,
\label{eq_sigmaXX_white}\eeq 
the above covariances simplify to 
\beq
\mathrm{cov}(\delta\Sigma^{xx})_{ij,kl} = \frac{1}{S} \left[ \sigma_x^4 \delta_{ik}\delta_{jl} + \delta_{il}\delta_{jk}\right] ,\qquad 
\mathrm{cov}(\delta\Sigma^{yx})_{ij,kl} = \frac{1}{S} \left[ \bra y_i^2\ket \sigma_x^2 \delta_{ik} \delta_{jl} + \Sigma_{0,il}^{yx}\Sigma_{0,jk}^{yx}\right].
\label{eq_cov_sigma_white}\eeq 

Upon inserting these expansions of the data covariance matrices into \cref{eq_2L_SGD} and proceeding to the continuum limit analogously to the single-layer case [\cref{eq_SGD_Langevin}], we obtain the following \emph{stochastic gradient flow} (SGF) equations:
\begin{subequations}
\begin{align} 
\pd_t W_1 &= W_2^T\left(\Sigma_0^{yx} - W_2 W_1 \Sigma_0^{xx}\right) + \Rcal_1, \\
\pd_t W_2 &= \left(\Sigma_0^{yx} - W_2 W_1 \Sigma_0^{xx}\right) W^T_1 + \Rcal_2,
\end{align}\label{eq_2L_SGF}
\end{subequations}
with the multiplicative noise matrices $\Rcal_1\in\reals^{N_h\times N_i}$, $\Rcal_2\in\reals^{N_o\times N_h}$:
\begin{subequations}
\begin{align}
\Rcal_1 &= \sqrt{\lambda} \, W_2^{T}  \big( \delta\Sigma^{yx} -  W_2 W_1 \delta\Sigma^{xx} \big), \\
\Rcal_2 &= \sqrt{\lambda} \big( \delta\Sigma^{yx}  - W_2 W_1 \delta\Sigma^{xx}  \big) W_1^{T} .
\end{align}\label{eq_2L_SGF_noises}
\end{subequations}

\subsubsection{Perturbation expansion}

In order to analyze the stationary fluctuation dynamics, we introduce a suitable linearization into \cref{eq_2L_SGF}.
Accordingly, we decompose the weight matrices as $W_{1,2} = W_{1,2}^0 + \delta W_{1,2}$, where $W_{1,2}^0$ denotes the solution of the full-batch GD and $\delta W_{1,2}$ describe small fluctuations around this solution.
Upon expanding \cref{eq_2L_SGF} up to linear order in the fluctuations of $W$ and $\Sigma$, and collecting terms of the same order, we obtain the deterministic gradient flow equations
\begin{subequations}
\begin{align}
\pd_t  W_1^0 &=  W^{0,T}_2 \left(  \Sigma_0^{yx} -  W_2^0 W_1^0 \Sigma_0^{xx} \right) , \\
\pd_t  W_2^0 &= \left( \Sigma_0^{yx} - W_2^0 W_1^0 \Sigma_0^{xx} \right) W_1^{0,T} ,
\end{align}\label{eq_2L_GF}
\end{subequations}
and the dynamics of the first order perturbations: 
\begin{subequations}
\begin{align}
\pd_t \delta W_1 &= \big( \delta W_2^{T} \Sigma_0^{yx}  - \delta W^T_2 w^* \Sigma_0^{xx} - W^{0,T}_2 \delta W_2 W_1^0 \Sigma_0^{xx} - W_2^{0,T} W_2^0 \delta W_1\Sigma_0^{xx}  \big) + \Rcal_1, \\
\pd_t \delta W_2 &= \big(  \Sigma_0^{yx} \delta W_1^T  - \delta W_2 W_1^0 \Sigma_0^{xx} W_1^{0,T}  - W_2^0 \delta W_1 \Sigma_0^{xx}  W_1^{0,T} - w^* \Sigma_0^{xx}\delta W_1^T \big) + \Rcal_2.
\end{align}\label{eq_2L_lindyn_full}
\end{subequations}
in which the noises [\cref{eq_2L_SGF_noises}] now become additive (and are assumed to be quantities of the same order as $\delta W_{1,2}$):
\begin{subequations}
\begin{align}
\Rcal_1 &= \sqrt{\lambda} \, W_2^{0,T}  \big( \delta\Sigma^{yx} -  W_2^0 W_1^0 \delta\Sigma^{xx} \big) =  \sqrt{\lambda}\, W_2^{0,T}  \big( \delta\Sigma^{yx} -  w^* \delta\Sigma^{xx} \big), \\
\Rcal_2 &= \sqrt{\lambda} \big( \delta\Sigma^{yx}  - W_2^0 W_1^0 \delta\Sigma^{xx}  \big) W_1^{0,T} = \sqrt{\lambda} \big( \delta\Sigma^{yx}  - w^* \delta\Sigma^{xx}  \big) W_1^{0,T}.
\end{align}\label{eq_2L_noise_expr}
\end{subequations}
Crucially, we focus henceforth on a (quasi-)\emph{stationary state} of training, where the $W_{1,2}^0$ are essentially time independent or at least change on a significantly longer time scale than the $\delta W_{1,2}$, allowing us to analyse \cref{eq_2L_lindyn_full,eq_2L_noise_expr} with $W_{1,2}^0\simeq \const$. 
Since explicit expressions for $W_i^0$ in terms of the data and initial conditions are not available, we will consider the $W_{1,2}^0$ as given outcomes of the gradient flow [\cref{eq_2L_GF}], obtained, e.g., from a numerical solution \footnote{Note that the zero-balanced solution in \cref{eq_2L_wtsol_zbal} involves different matrix products of the $W_i$ from the ones in \cref{eq_2L_SVD_W12} and thus cannot be directly used here.}.

For whitened inputs [see \cref{eq_sigmaXX_white}], we have $w^*=\Sigma_0^{yx} / \sigma_x^2 $ [see \cref{eq_LR_sol}] as well as
\beq (\Sigma_0^{yx} \Sigma_0^{yx,T})_{ij} = \frac{1}{P^2} \sum_{\mu,\nu}\sum_k y_i^\mu x_k^\mu x_k^\nu y_j^\nu \simeq \frac{1}{P^2} \sum_\mu (y^\mu)^2\sigma_x^2 \delta_{ij} = \frac{1}{P}\sigma_x^2 \bra y^2\ket \delta_{ij},
\label{eq_sigmaYX_squared}\eeq 
where we used $\xv^\mu\cdot \xv^\nu \simeq \sigma_x^2\delta^{\mu\nu}$ and the fact that $\yv^\mu$ are i.i.d.\ variates. 
With these simplifications, the noise covariance takes the following form (see \cref{app_noise_cov}):
\begin{subequations}
\begin{align}
\cov(\Rcal_1)_{ij,kl} &= \sigma_\Rcal^2 \delta_{jl} \left( 1-\sfrac{1}{P}\right) (W_2^{0,T} W_2^0)_{ik},\qquad j,l=1,\ldots,N_i, \label{eq_2L_noisecov_R1} \\
\cov(\Rcal_2)_{ij,kl} &= \sigma_\Rcal^2 \delta_{ik} \left(1-\sfrac{1}{P}\right) (W_1^0 W_1^{0,T})_{jl}, \qquad i,k=1,\ldots,N_o, \label{eq_2L_noisecov_N2} \\
\cov(\Rcal_1,\Rcal_2)_{ij,kl} &= \sigma_\Rcal^2 (1-1/P) W^{0,T}_{2,ik} W^{0,T}_{1,jl} \qquad \text{with}\qquad \sigma_\Rcal^2\equiv \frac{1}{S} \lambda \sigma_x^2  \bra y^2\ket,
\end{align} \label{eq_2L_noisecov}
\end{subequations}
while \cref{eq_2L_lindyn_full} becomes
\begin{subequations}
\begin{align}
\pd_t \delta W_1 &=  -\sigma_x^2 \big[ W_2^{0,T} W_2^0 \delta W_1 + W^{0,T}_2 \delta W_2 W_1^0  \big] + \Rcal_1 = -\frac{\pd \tilde L}{\pd \delta W_1} + \Rcal_1, \\
\pd_t \delta W_2 &=  -\sigma_x^2 \big[ \delta W_2 W_1^0 W_1^{0,T} + W_2^0 \delta W_1  W_1^{0,T} \big] + \Rcal_2 = -\frac{\pd \tilde L}{\pd \delta W_2} + \Rcal_2,
\end{align}\label{eq_2L_lindyn}
\end{subequations}
\hspace{-0.1cm}with the reduced loss $\tilde L(\delta W_1, \delta W_2)\equiv \onehalf \sigma_x^2\, \tr \left[ W_2^{0,T} W_2^0 \delta W_1 \delta W_1^T + \delta W_2^{0,T} W_2^0 \delta W_1 W_1^{0,T} + \delta W_2^T \delta W_2 W_1^0 W_1^{0,T} \right] $ quadratic in the fluctuations.

Recalling the matrix rank equation $\rank(AB) = \min(\rank A, \rank B)$, we note that, for one-dimensional outputs ($N_o=1$), one has $\rank(W_2^0)=1=\rank(W_2^{0,T} W_2^0)$ and thus $\rank([\cov(\Rcal_1)])=N_i$ according to \cref{eq_2L_noisecov_R1}.
For zero-balanced initial conditions [i.e.\ \cref{eq_2L_wtcons} with vanishing r.h.s.], one has $\text{rank}( [\cov(\Rcal_2)])=\text{rank}(W_1^0 W_1^{0,T})=\text{rank}(W_2^{0,T} W_2^0)=1$, while for random initial conditions, typically $\text{rank}([\cov(\Rcal_2)]) = \text{rank}(W_1^0 W_1^{0,T})= N_h$.

\subsubsection{Transformations}
\label{sec_2L_trafos}

In order to further analyze \cref{eq_2L_lindyn}, we expand $\delta W_{1,2}$ in the eigenbasis of $W_{1,2}^0$. To this end, we introduce the thin SVD $W_1^0 = A S_1 B^T$ [where $A$, $S_1$ and $B$ are $(N_h,N_h), (N_h,N_i)$ and $(N_i,N_i)$ matrices], and the standard SVD $W_2^0 = U S_2 V^T$ [where $U$, $S_2$ and $V$ are $(N_o,N_o), (N_o,N_o)$ and $(N_h,N_o)$ matrices], which yield 
\begin{subequations}
\begin{align}
 W_1^0 W_1^{0,T} &= A D_1 A^T\quad \text{with}\quad  D_1\equiv S_1 S_1^T \in \reals^{N_h\times N_h}, \\ W_2^{0,T} W_2^{0} &= V D_2 V^T \quad \text{with}\quad  D_2\equiv S_2^T S_2\in \reals^{N_o\times N_o}.
\end{align} \label{eq_2L_SVD_W12}
\end{subequations}
\hspace{-0.2cm}The eigenvalues of $W_1^0 W_1^{0,T}$ and $W_2^{0,T} W_2^0$ encoded in the diagonal positive-semidefinite matrices $D_{1,2}$ will play a crucial role in the following. 
Introducing new dynamical variables $z_1 \equiv V^T \delta W_1\in \reals^{N_o\times N_i}$, $z_2 \equiv \delta W_2 A \in \reals^{N_o\times N_h}$, and noises $\eta_1 \equiv V^T \Rcal_1\in \reals^{N_o\times N_i}$, $\eta_2 \equiv \Rcal_2 A\in \reals^{N_o\times N_h}$, reduces \cref{eq_2L_lindyn} to
\begin{subequations}
\begin{align}
\dot z_1 &= -D_2 z_1 - (US_2)^T z_2 S_1 B^T + \eta_1, \\ 
\dot z_2 &= -z_2 D_1 - US_2\, z_1 B S_1^T + \eta_2.
\end{align}\label{eq_2L_projdyn}
\end{subequations}
The covariance of the new noises $\eta_{1,2}$ follows readily by using the SVDs in \cref{eq_2L_noisecov}:
\begin{subequations}
\begin{align}
\cov(\eta_1)_{mj,nl} &= \sigma_\Rcal^2 \delta_{jl} D_{2,mn}, \qquad (j,l=1,\ldots N_i,\quad m,n=1,\ldots,N_o), \\
\cov(\eta_2)_{im,kn} &= \sigma_\Rcal^2 \delta_{ik} D_{1,mn}, \qquad (m,n=1,\ldots N_h,\quad i,k=1,\ldots,N_o),\\
\cov(\eta_1,\eta_2)_{mj,kn} = \cov(\eta_2,\eta_1)_{nk,jm} &= \sigma_\Rcal^2 (US_2)_{mk}^T (B S_1^T)_{jn}, \qquad (j=1,\ldots,N_i,\quad n=1,\ldots,N_h, \quad m,k=1,\ldots,N_o).
\end{align} \label{eq_2L_projnoise_cov}
\end{subequations}
Since the $D_i$ are diagonal, these results represent a significant simplification over \cref{eq_2L_lindyn,eq_2L_noisecov}.

We now introduce the joint vectors $\Zv\equiv \{\vect(z_1), \vect(z_2)\} \in \reals^{N_o(N_i+N_h)}$, $\boldsymbol{\eta}\equiv \{\vect(\eta_1),\vect(\eta_2)\}$, which allows us to express \cref{eq_2L_projdyn} in the compact form 
\beq \dot \Zv = -\Gamma \Zv + \boldsymbol{\eta}.
\label{eq_2L_vecdyn}\eeq 
The drift matrix $\Gamma$ and the diffusion matrix $\Delta \equiv \bra \boldsymbol{\eta}\boldsymbol{\eta}^T\ket$ turn out to be identical symmetric square matrices of size $[N_o(N_i+N_h)]^{2}$: 
\beq \Gamma / \sigma_x^2 = \begin{pmatrix}
 \Imat_{N_i}\otimes D_2  & B S_1^T \otimes (U S_2)^T  \\
  S_1 B^T \otimes US_2 &  D_1\otimes \Imat_{N_o}
\end{pmatrix} = \Delta/\sigma_\Rcal^2 .
\label{eq_2L_drift_diff}\eeq 
Here,  $\otimes$ denotes the Kronecker product and in the derivation, we used $\vect(\Acal\Bcal\Ccal) = (\Ccal^\mathrm{T}\otimes \Acal) \vect(\Bcal)$ as well as $\vect(\Acal \Bcal) = (\Imat_{\mathrm{dim}_2 \Bcal} \otimes \Acal) \vect(\Bcal) = (\Bcal^\mathrm{T}\otimes \Imat_{\mathrm{dim}_1\Acal}) \vect(\Acal)$ where $\mathrm{dim}_n \Acal$ denotes the size of the $n$th dimension of $\Acal$ \cite{petersen_matrix_2012}.
Next, we determine the eigenvalues and the rank of $\Gamma/\sigma_x^2$ and $\Delta/\sigma_\Rcal^2$. After some algebra, using the determinant formula for block matrices \cite{petersen_matrix_2012} and assuming $N_i\geq N_h\geq N_o$, we find the eigenvalues \footnote{This notation means that, e.g., in \cref{eq_2L_eval_gamma1}, all possible combinations of $k$ and $l$ are to be taken, resulting in $N_h N_o$ distinct eigenvalues, while in \cref{eq_2L_eval_gamma2}, each $D_{2,ll}$ occurs with multiplicity $N_i-N_h$.} 
\begin{subequations}
\begin{align}
\text{multiplicity}\, 1: &\quad  D_{1,kk}+D_{2,ll},\qquad \text{with}\quad k=1,\ldots,N_h,\quad l=1,\ldots,N_o, \label{eq_2L_eval_gamma1} \\
\text{multiplicity}\, N_i-N_h: &\quad  D_{2,ll},\qquad \text{with}\quad l=1,\ldots,N_o ,\label{eq_2L_eval_gamma2} \\
\text{multiplicity}\, N_h N_o: &\quad 0.
\end{align}\label{eq_2L_eval_gamma}
\end{subequations}
Accordingly, $\Gamma$ and $\Delta$ are positive semi-definite matrices with $\rank(\Gamma)=\rank(\Delta)=N_o N_i$. The rank deficiency of $\Gamma$ and $\Delta$ turns out to be a crucial property in the following.

\subsubsection{General properties}
\label{sec_2L_genprop}
From the results of the preceding subsection one can already infer a few general properties of the fluctuations $\delta W_{1,2}$, which become particularly simple in the case $N_o=1$: First, from the definition $\delta W_1(t)=V z_1(t) = (V z_{1,1}(t), V z_{1,2}(t), \ldots, V z_{1,N_i}(t))$, with the column vector $V$, we infer that all $N_h$ elements in the $i$th column of $\delta W_1$ fluctuate in the same way, i.e., their dynamics is controlled by the scalar $z_{1,i}(t)$, up to time-independent prefactors contained in $V$. This applies independently from the initialization. 
Next, we recall $\delta W_2(t) = z_2(t) A^T = (z_2(t)\cdot \bar a_1, z_2(t)\cdot \bar a_2, \ldots, z_2(t)\cdot \bar a_{N_h})$, where $\bar a_i$ denotes the $i$th column of $A^T$. In the case of zero-balanced initial conditions [described by \cref{eq_2L_wtcons} with vanishing r.h.s.], one has $1=\rank(W_2^0)=\rank(W_2^{0,T} W_2^0) = \rank(W_1^0 W_1^{0,T})$ and hence $D_1=\diag(r, 0,\ldots,0)$ [see \cref{eq_2L_SVD_W12}]. Accordingly, in this case, the scalars $z_{2,i}(t) = 0$ for $i>1$, such that $\delta W_2(t) = (\bar a_{1,1}, \bar a_{2,1},\cdots, \bar a_{N_h,1}) z_{2,1}(t)$, i.e., all elements of $\delta W_2$ fluctuate identically up to individual scaling factors determining their variance. By contrast, in the case of random initialization, $D_1$ is typically of maximum rank, implying that all elements of $\delta W_2$ fluctuate independently. 
Numerical results show that the above statements apply also for non-white $\Sigma^{xx}$.

\subsection{Steady-state fluctuations of weights}

Having obtained the linear Langevin equation [\cref{eq_2L_vecdyn}] for the weights in each layer, we proceed to determine the resulting steady-state statistics of the weight fluctuations. We find that, as a genuine consequence of the structure of a deep linear network, weight fluctuations are generally anisotropic. We finally provide a derivation of the IVFR.

\subsubsection{Covariances}

\begin{figure}[b]\centering
   \subfigure[]{\includegraphics[width=0.27\linewidth]{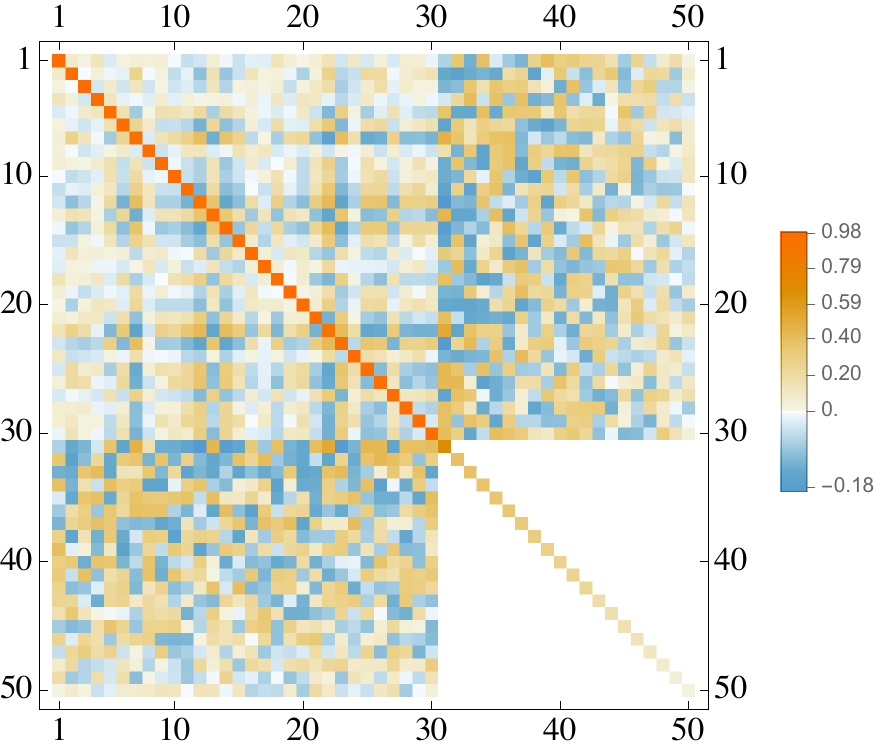} \label{fig_2L_covar_mat}}\quad
   \subfigure[]{\includegraphics[width=0.32\linewidth]{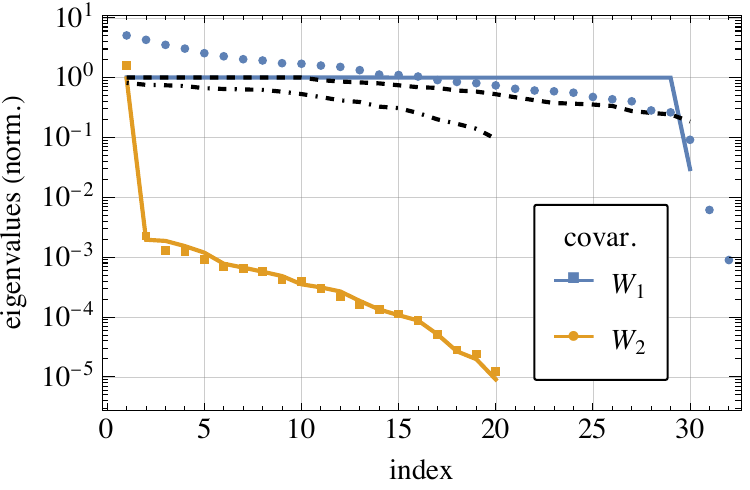} \label{fig_2L_cov_datadom}}\quad
   \subfigure[]{\includegraphics[width=0.32\linewidth]{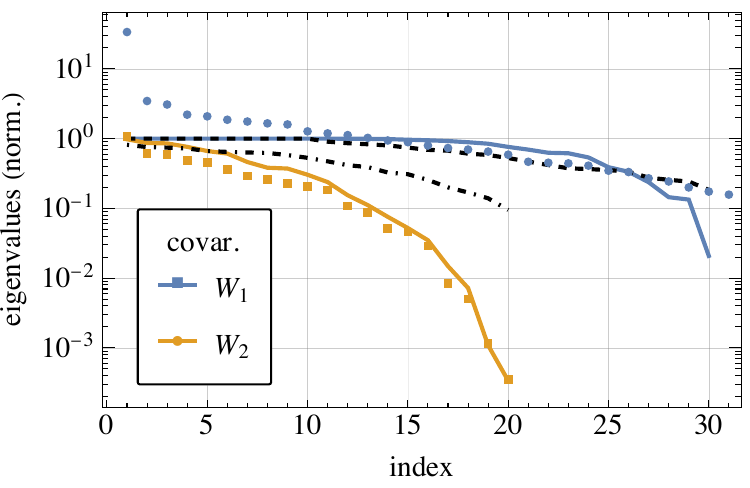} \label{fig_2L_cov_initdom}}\quad 
   \caption{Covariance of the weights of a two-layer linear network in the quasi-stationary state. (a) Visualization of the covariance matrix $M$ [\cref{eq_2L_varZ}] of the transformed weights $z_i$ for a typical simulation with $N_i=30$, $N_h=20$, $N_o=1$. In the special basis defined by the SVD of $W_1^0$ and $W_2^0$ [see \cref{eq_2L_SVD_W12}], the covariance of the second-layer weights is fully diagonal. (b,c) Eigenvalues of $M$ obtained from SGD experiments (symbols), compared to the theoretical predictions obtained from $M_{11}$ and $M_{22}$ in \cref{eq_2L_varZ} (solid lines), for a single realization of the data. In the simulations, we use non-white Gaussian i.i.d.\ samples $x^\mu_i\sim \Ncal(0,1/N_i)$, label noise variance of $\sigma_y^2=0.1$, learning rate $\lambda=0.1$, mini-batch size $S=10$, and an initial variance of the weights of (b) $\sigma_{W_1}^2 = 10^{-4}/N_i$, $\sigma_{W_2}^2 = 10^{-4}/N_h$, and (c) $\sigma_{W_1}^2 = 1/N_i$, $\sigma_{W_2}^2 = 1/N_h$. For comparison, we show the eigenvalues of $M$ obtained by assuming Gaussian i.i.d.\ distributed elements for $W_{1}^0$ and $W_2^0$ as dashed and dot-dashed lines, respectively. Eigenvalues are normalized by $\lambda \sigma_y^2/(2S)$ [cf.\ \cref{eq_2L_covarW2}]. }
   \label{fig_2L_var}
\end{figure}

The covariance matrix $M\equiv \cov(\Zv) = \bra \Zv \Zv^T\ket$ describing the fluctuations of the $z_i$ is determined by a Lyapunov equation \cite{risken_fokkerplanck_1989, laub_matrix_2005, mandt_stochastic_2017, kunin_limiting_2021}:
\beq \Gamma M + M \Gamma^T = \Delta.  
\label{eq_Lyapunov}\eeq 
As one readily checks (using the fact that $\Gamma\propto \Delta$ and the identity $\Acal \Acal^+ \Acal=\Acal$ for a matrix $\Acal$), \cref{eq_Lyapunov} is solved by  
\beq M \equiv \begin{pmatrix}
	M_{11} & M_{12} \\
	M_{21} & M_{22}
\end{pmatrix} = \frac{1}{2} \Gamma^+ \Delta,
\label{eq_2L_varZ}\eeq 
which is symmetric of size $N_o (N_i+N_h)\times N_o(N_i+N_h)$, 
while $M_{11}\in\reals^{N_o N_i \times N_o N_i}$, $M_{12},M_{21}^T\in\reals^{N_o N_i\times N_o N_h}$, $M_{22}\in\reals^{N_o N_h\times N_o N_h}$.
This solution assumes detailed balance, which is approximately realized for SGD in the case of sufficient underparameterization [see \cref{eq_wts_cov_detbal} and the related discussion]. 
The covariances of the weight fluctuations $\delta W_1$ and $\delta W_2$ in the original space follow as
\begin{subequations}
\begin{align}
\cov(\vect(\delta W_1)) &= (\Imat_{N_i}\otimes V) M_{11} (\Imat_{N_i}\otimes V^T), \label{eq_2L_covW1} \\
\cov(\vect(\delta W_2)) &= (A\otimes \Imat_{N_o}) M_{22} (A^T \otimes \Imat_{N_o}), \label{eq_2L_covW2} 
\end{align}
\label{eq_2L_covW12} 
\end{subequations}
with analogous relations for the cross-covariances. Since $(\Acal\otimes\Bcal)^T=\Acal^T\otimes B^T$], the eigenvalues of $M_{22}$ coincide with the eigenvalues of $\cov(\vect(\delta W_2))$, while the eigenvalues of $M_{11}$ coincide with the nonzero eigenvalues of $\cov(\vect(\delta W_1))$, noting that the latter has at least $N_i (N_h-N_o)$ vanishing eigenvalues. 
A typical covariance matrix $M$ is visualized in \cref{fig_2L_covar_mat}.
Upon averaging over the data distribution, all off-diagonal elements vanish, leaving only the diagonal ones as ``universal'' imprints of the data statistics (see also the discussion below).
However, since we are interested in properties pertaining to a specific realization of the data, we do not perform such averaging here.

Due to the rank deficiency of $\Gamma$, the matrix $M$ is, in general, non-diagonal. This can also be inferred from the eigen-decompositions $\Gamma/\sigma_x^2=\Delta/\sigma_\Rcal^2 = R \Omega R^T$, where $\Omega$ is a diagonal matrix of rank $N_o N_i$ [see \cref{eq_2L_eval_gamma}], which implies $M = (R^{T})^{-1} \Omega^+ \Omega R^T$ with a diagonal matrix of eigenvalues 
\beq \Omega^+ \Omega = \frac{\lambda \sigma_y^2}{2S} \Imat_{N_o(N_i+N_h)}^{N_o N_i} . 
\label{eq_2L_cov_eval}\eeq 
Accordingly, $M$ is of rank $N_o N_i$ and has $N_o N_h$ vanishing eigenvalues. 

In order to limit complexity of the analysis, we focus in the following on a one-dimensional output $N_o=1$. In this case, $z_i$ and $\eta_i$ become row vectors, while $U$, $S_2$, $D_2$ become scalars (in fact,  $U=\pm 1$), and
\beq 
\Gamma/\sigma_x^2 = \Delta/\sigma_\Rcal^2 =  \begin{pmatrix}
D_2  \Imat_{N_i} & (US_2)  B S_1^T \\
(U S_2)  S_1 B^T &  D_1
\end{pmatrix} 
= \begin{pmatrix}
V^T W_2^{0,T} W_2^{0} V &  (W_2^{0}V)^T W_1^{0,T} A \\
(W_2^{0}V) A^T W_1^0 &  A^T W_1^0 W_1^{0,T} A 
\end{pmatrix} .
\label{eq_2L_GammaDelta_1d}\eeq 
Using expressions for the pseudo-inverse of a block matrix \cite{miao_general_1991}, we find, after some tedious but straightforward algebra, that $M_{22}$ [\cref{eq_2L_varZ}] has a simple diagonal form:
\beq M_{22} = \cov(z_2) = \frac{\lambda\sigma_y^2}{2S} \, \mathrm{diag} \left\{\frac{D_{1,ii}}{D_{1,ii} + D_2} \right\}_{i=1,\ldots,N_h}.
\label{eq_2L_covarW2}\eeq 
Accordingly, the components of $z_2$ [see \cref{eq_2L_projdyn}] fluctuate independently and $M_{22,ii}$ in fact represent the eigenvalues of $\cov(\delta W_2)$ [see \cref{eq_2L_covW2}].
By contrast, $M_{11}$ is, in general, not diagonal [see \cref{fig_2L_covar_mat}], although its off-diagonal elements are typically small, implying that the components of $z_1$ are weakly cross-correlated.

\subsubsection{Covariances: discussion and numerical results}
\label{sec_2L_discuss}

We conclude this analysis with several remarks: (1) Without the cross coupling between $\delta W_1$ and $\delta W_2$ [see \cref{eq_2L_lindyn}], $\Gamma$ and $\Delta$ would be diagonal and full rank, which would imply $M\propto \Imat_{N_i+N_h}$ and thus $[\cov(\delta W_1)] \propto \Imat_{N_i+N_h}$ and $[\cov(\delta W_2)]\propto \Imat_{N_o\times N_h}$, i.e.\ isotropic fluctuations. 
(2) In the case of Gaussian i.i.d.\ data considered here, the analysis starting from \cref{eq_2L_SVD_W12} essentially relied only on the eigenstructure of the weight product matrices $W_1^0 W_1^{0,T}$ and $W_2^{0,T}W_2^0$, which enter both the drift and the diffusion matrix [\cref{eq_2L_drift_diff}]. The data is assumed to be structureless [with whitened inputs, see \cref{eq_sigmaXX_white}] and thus appears here only as scaling factors $\sigma_x^2$, $\sigma_y^2$.
Numerical results confirm that similar fluctuation spectra are also obtained for non-white data.
(3) The solution in \cref{eq_2L_LRsol} only determines the $N\eff<N\st{tot}$ degrees of freedom of $W_2^0 W_1^0$ uniquely in terms of the data, implying that the individual weight matrices $W_1^0$ and $W_2^0$ (and hence also $D_{1,2}$) are additionally affected by the initialization. 
Specifically, in the limit $N\st{tot}\gg N\eff$, the weights hardly change from their initial values during training (lazy training regime) \cite{chizat_lazy_2019, geiger_landscape_2021}, allowing one to approximate a nonlinear neural network by its linearization around initialization \cite{jacot_neural_2018, lee_wide_2019}. Assuming $W_i^0(t=0)\sim \Ncal(0,\sigma_{W_i})$, $W_1^0 W_1^{0,T}$ and $W_2^{0,T}W_2^0$ in this case can be approximated as Wishart matrices, the eigenvalues of which asymptotically follow a Marchenko-Pastur distribution with characteristic magnitude $D_{i}\propto \sigma_{W_i}^2$. 
When initializing the weights conventionally with \cite{glorot_understanding_2010}
\beq \sigma_{W_1}^2\simeq 1/N_i, \qquad \sigma_{W_2}^2\simeq 1/N_h,
\label{eq_convent_init}\eeq  
and assuming $N_h\lesssim N_i$, one accordingly expects $D_{1,ii}\lesssim D_2$ [see \cref{eq_2L_SVD_W12}], such that \cref{eq_2L_covarW2} can be approximated as
\beq M_{22}\simeq \frac{\lambda\sigma_y^2}{2 S D_2}D_1,
\label{eq_2L_covarW2_approx}\eeq 
except for the few largest eigenvalues in $D_1$. 
We conclude that for deep linear nets initialized in this way, weight fluctuations are generally \emph{anisotropic} and their spectrum has a certain universal character encoded in the limiting eigenvalue distribution.
A notable exception occurs for zero-balanced initial conditions [see \cref{eq_2L_wtcons}], where, owing to $D_{1,ii}\propto \delta_{i,0}$ [see \cref{sec_2L_genprop}], the fluctuation spectrum is isotropic.
(4) We recall that the derivation leading to \cref{eq_2L_covarW2,eq_2L_covarW2_approx}, demonstrating the anisotropy of the fluctuation spectrum, is obtained within the large-sample number limit [$P\to\infty$; see \cref{eq_2L_cov_largeP} and the related discussion], which, in the case of a single-layer net, would imply isotropic weight fluctuations [see \cref{eq_wts_cov_conv}]. 

\Cref{fig_2L_var}(b,c) shows simulation results obtained from SGD training of a two layer linear network on (non-white) Gaussian i.i.d.\ data for Gaussian weight initializations [\cref{eq_convent_init}] of (b) small and (c) large variance. 
We find good agreement with the covariance spectrum of the weights predicted by \cref{eq_2L_varZ}, which is a linearization of the full stochastic gradient flow [see \cref{eq_2L_lindyn_full}]. 
In particular, the assumption of whitened data in the theory does not appear to qualitatively affect the results.
Confirming \cref{eq_2L_covarW2}, the spectrum of the fluctuations of $W_2$ is essentially determined by the eigenvalues of $W_1^0 W_1^{0,T}$ [see \cref{eq_2L_SVD_W12}].
Eigenvalues of $\cov(\delta W_1)$ with index larger than $N_i$ vanish according to the linear theory [\cref{eq_2L_varZ,eq_2L_covW12}], while in our simulations, they are nonzero but decrease in magnitude with increasing index.
Comparing the two panels, we observe that, for large values of the initial weight variance, the spectrum remains closer to the shape implied for Gaussian i.i.d.\ elements of $W_{1,2}^0$.
For weights close to being zero-balanced, the fluctuations of $W_2$ are essentially governed by a single independent mode [see discussion in \cref{sec_2L_genprop}], which is reflected by the prominent peak of the lowest mode in \cref{fig_2L_var}(b). The eigenvalue peak of the covariance of $W_1$ seen in (c) can be associated to an overall drift of the mean of $W_1$, which occurs on a large time scale. 
In fact, although the loss is stationary, we observe in our SGD experiments a transient dynamics of the individual weights, resulting in a slow but steady drop of the overall magnitude of the eigenvalues (except for the zero mode).
This dynamics, which is most pronounced for small batch sizes, is not captured by the present analysis, since the latter only provides a relation between the covariances of $W_{1,2}$ and the products $W_1^0 W_1^{0,T}$, $W_2^{0,T}W_2^0$, \emph{assuming} quasi-stationary means $W_1^0$ and $W_2^0$. A complete description requires a solution of the GF equations \eqref{eq_2L_GF} and is reserved for a future study.

\subsubsection{Inverse variance-flatness relation}

\begin{figure}[t]\centering
   \subfigure[]{\includegraphics[width=0.36\linewidth]{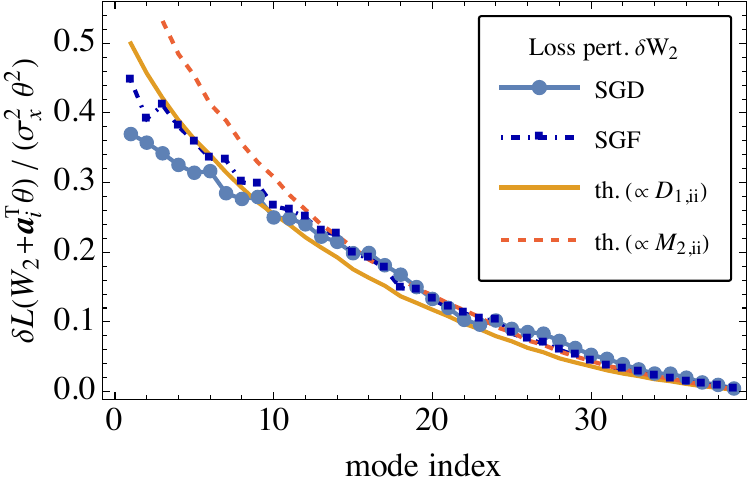} }\qquad 
   \subfigure[]{\includegraphics[width=0.37\linewidth]{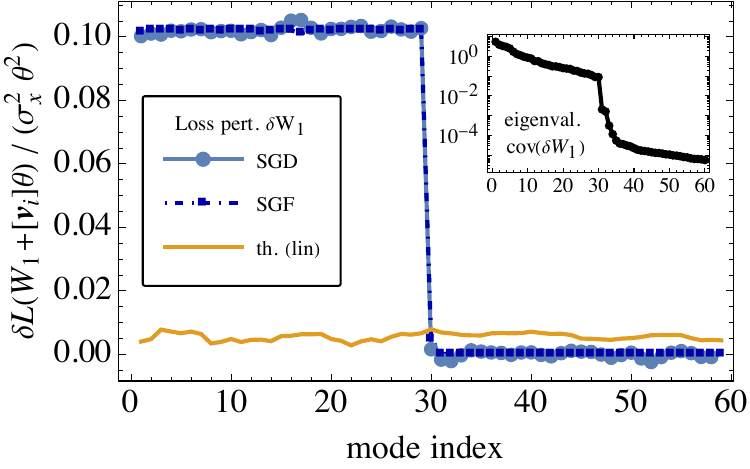} }
   \caption{Perturbing the loss $L$ of a two-layer linear network [\cref{eq_2L_loss}] by the eigenvectors of the covariance matrix of the fluctuations of (a) $\delta W_2$ [see \cref{eq_2L_losspert}] and (b) $\delta W_1$. Results of training on Gaussian i.i.d.\ data ($\sigma_x^2=1/N_i$, $\sigma_y^2=1$) with SGD (connected circles) are compared to the theoretical predictions (solid lines), given by \cref{eq_2L_losspert} in (a), and by the numerically evaluated $\delta L(\theta \vv_i)$ using the eigenvectors $\vv_i$ of $\cov(\vect(\delta W_1))$ [\cref{eq_2L_covW1}] in (b). The dashed-dotted line represents the prediction based on the nonlinear stochastic gradient flow.  In (a), beside the exact result $\delta L=\sigma_x^2 \theta^2 D_{1,ii}/2$ (solid line), also the approximation [see \cref{eq_2L_losspert}] in terms of the eigenvalues of the covariance $M_{22}$ is plotted (dashed line). The inset in (b) shows the eigenvalues of $\cov(\vect(\delta W_1))$ ordered by magnitude vs.\ the mode index $i$. For $i>N_i$, the eigenvalues abruptly drop to a smaller level, while they are predicted to vanish according to the linear model (see discussion in \cref{sec_2L_discuss}). The corresponding drop of $\delta L$ at $i=N_i$ is captured by the nonlinear stochastic gradient flow simulations, while the linear theory [solid line in (b)] instead predicts a flat profile for $\delta L$.  
   We used $N_o=1$, $N_h=40$, $N_i=50$ in (a) and $N_o=1$, $N_h=20$, $N_i=30$ in (b). The behaviors for whitened and non-whitened data are similar.
   }
   \label{fig_2L_losspert}
\end{figure}

We now study how the fluctuation variance relates to the loss, thereby providing an analytical derivation of the IVFR \cite{feng_inverse_2021}.
The reduced quadratic loss [see \cref{eq_2L_lindyn}] is given by $\tilde L(z_1,z_2) = \onehalf \sigma_x^2 \tr [ D_2 z_1 z_1^T + z_2^T z_2 D_1 + z_2^T (US_2) z_1 (BS_1^T)]$, which indicates that the perturbation of the loss due to fluctuations of $z_1$ and $z_2$ is controlled by the spectrum of $W_2^{0,T}W_2^0$ and $W_1^0 W_1^{0,T}$, respectively [see \cref{eq_2L_SVD_W12}].
In order to capture the effect of the fluctuations in the original weight space, we start from \cref{eq_2L_loss} and recall that the principal modes $\av_{i=1,\ldots,N_h}$ of the fluctuations of $W_2$ (i.e., the columns of $A$) fulfill $W_1^0 W_1^{0,T} \av_i = D_{1,ii} \av_i$ with $ D_{1,ii} \simeq 2 S D_2 M_{22,ii}/(\lambda\sigma_y^2)$ [see \cref{eq_2L_covW12,eq_2L_covarW2_approx}].
Using the basic relation $\tr(\Acal\Bcal)=\tr(\Bcal\Acal)$ as well as the fact that $W_2^0 W_1^0 = \sigma_x^{-2} \Sigma_0^{yx}$ for whitened inputs, the perturbation of the loss [\cref{eq_2L_loss}] along $\av_{i}$ follows, after a bit of algebra, as [cf.\ \cref{eq_loss_pert_def}]: 
\begin{equation}
\delta L(\theta\av_i) = L(W_1^0, W_2^0+ \theta \av_i^T) - L(W_1^0,W_2^0)  = \frac{1}{2}\sigma_x^2 \theta^2 D_{1,ii} \simeq \frac{S D_2\sigma_x^2}{\lambda\sigma_y^2} \theta^2  M_{22,ii} \,.\label{eq_2L_losspert}\end{equation}
As $W_2^0\in \reals^{N_o\times N_h}$ and $N_o=1$, we reshaped the vector $\av_i$ into a compatible form via its transpose.
Numerical experiments with SGD in the large batch regime as well as with SGF agree well with this prediction [see \cref{fig_2L_losspert}(a)].
Since the ``flatness'' $F_i$ of the loss $\propto$ (curvature $\kappa_i$)$^{-1/2}$ and the curvature $\kappa_i$ of the parabolas defined by \cref{eq_2L_losspert} is given by their quadratic coefficient, i.e., $\kappa_i = 2 S\sigma_x^2 D_2 M_{22,ii}/(\lambda\sigma_y^2)$, we recover the inverse variance flatness relation in the form $M_{22,ii}\propto F_i^{-\psi}$ with $\psi =2$. In Ref.\ \cite{feng_inverse_2021}, empirically $\psi\approx 4$ was found in the case of cross-entropy loss, but it was emphasized that the value of the exponent can depend on the specific training setting.
We remark that, due to the transient weight dynamics persisting under SGD (see \cref{sec_2L_discuss}), we observe that $\delta L$ slowly diminishes with time, while remaining consistent with the IVFR.

By contrast, for a perturbation of the loss along eigendirections $\vv_i$ of $\cov(\vect\,\delta W_1)$ [\cref{eq_2L_covW1}], we do not observe any dependence on the mode index $i$, i.e., $\delta L(\theta\vv_i)=L(W_1^0+\theta [\vv_i],W_2^0)-L(W_1^0, W_2^0)\simeq \const.$, even though the weight covariance is anisotropic [see \cref{fig_2L_losspert}(b)]. Here, $[\vv_i]$ represents the vector $\vv_i$ reshaped into the form of $W_1^0$. Note, however, that $\delta L$ determined within the linearized fluctuation theory [i.e., using \cref{eq_2L_varZ}] does not capture the bimodal structure of $\delta L$ observed in numerical solutions of SGD or SGF [\cref{eq_2L_SGF}], but instead remains flat. We thus find that, for a two-layer linear network, the IVFR is not fulfilled for perturbations of the first-layer weights $W_1$. 

\subsection{Relaxation dynamics}

We illustrate here how the linearized SGF derived in the previous section can be used to analyze the  layer-wise relaxation dynamics of the weights.
Focusing on $N_o=1$, we diagonalize the (symmetric) drift and diffusion matrices as $\Gamma/\sigma_x^2 = \Delta/\sigma_\Rcal^2 =  R \Omega R^T$ [see \cref{eq_2L_GammaDelta_1d}], with a diagonal $\Omega=\diag(\omega_k)$ ($k=1,\ldots,N_h +N_i)$ and an orthogonal matrix $R$ with eigenvectors in the columns. According to \cref{eq_2L_eval_gamma}, the eigenvalues are given by
\beq \omega_{k=1,\ldots,N_h} = D_{1,kk}+D_2,\qquad \omega_{k=N_h+1,\ldots,N_i} = D_2,\qquad \omega_{k=N_i+1,\ldots,N_i+N_h}=0,
\eeq 
with $D_{1,2}$ defined in \cref{eq_2L_SVD_W12}. Note that all $\omega_k\geq 0$.
Introducing new dynamical variables $\boldsymbol\chi = R^T \Zv$ and noises $\boldsymbol\zeta = R^T \boldsymbol\eta$, we transform  \cref{eq_2L_vecdyn} to
\beq \dot \chi_k(t) = \sigma_x^2\omega_k \chi_k(t) + \zeta_k(t), \qquad \bra \zeta_k(t)\zeta_{k'}(t')\ket = \sigma_\Rcal^2\, \omega_k\,\delta_{kk'}\, \delta(t-t').
\eeq 
The general solution is given by $\chi_k(t) = e^{-\omega_k (t-t_0)} \chi_k(t_0) + \int_{t_0}^t \d s\, e^{\omega_k(s-t)}\zeta_k(s)$ with some initial time $t_0$.
Accordingly, in the stationary state, one obtains the correlation functions
\begin{align}
\bra\chi_k(t)\chi_{k'}(0)\ket &= \bra\chi_k(0)\chi_{k'}(0)\ket e^{-\sigma_x^2\omega_k t} ,\label{eq_2L_mode_relax} \\
\bra \chi_k(t)\chi_{k'}(t)\ket &= e^{-\sigma_x^2(\omega_k+\omega_{k'}) t} \bra\chi_k(0)\chi_{k'}(0)\ket + \delta_{kk'}\frac{\sigma_\Rcal^2}{2\sigma_x^2} \left(1-e^{-2\sigma_x^2\omega_k t}\right)  ,\label{eq_2L_mode_vardyn}
\end{align} 
where we assumed time-translation invariance, which allows us to chose the arbitrary time origin as $t_0=0$.
Modes $\chi_k$ belonging to eigenvalues $\omega_k=0$ remain fixed at their initialization $\chi_k(t)=\chi_k(0)$. By contrast, modes with $\omega_k>0$ relax towards zero [\cref{eq_2L_mode_relax}], while their variance saturates at late times at the equilibrium value [\cref{eq_2L_mode_vardyn}]:
\beq 
\bra \chi_k \chi_{k'}\ket\eq  = \begin{cases}
	\bra \chi_k(0) \chi_{k'}(0)\ket, \qquad & \omega_k=0=\omega_{k'}\\
	\delta_{kk'} \frac{\sigma_\Rcal^2}{2\sigma_x^2}, \qquad &\text{otherwise}. \\
\end{cases}
\eeq 
Notably, as a consequence of the equality of $\Gamma$ and $\Delta$ (up to an overall scale), the fluctuations of the modes $\chi$ are isotropic (independent of $k$) at late times and only determined by the relative strength of the noise ($\sigma_\Rcal^2$) and the loss ($\sigma_x^2$).
Since the initial values of $W_{1,2}$ are uncorrelated, we may assume $\bra \chi_k(0)\chi_{k'}(0)\ket \propto \delta_{kk'}$.
In order to transform back into the original weight space, one uses $\delta W_1 = V [\Pcal_1 R \boldsymbol\chi]^T$ and $\delta W_2 = [\Pcal_2 R\boldsymbol\chi]^T A^T$ [see \cref{eq_2L_SVD_W12} and notes that $\Zv$ is a column vector], with the projection matrices $\Pcal_1 = \begin{pmatrix}\Imat_{N_i} & 0 \\ 0 & 0 \end{pmatrix}$ and $\Pcal_2 = \begin{pmatrix}0 & 0 \\ 0 & \Imat_{N_h} \end{pmatrix}$, both of size $(N_i+N_h)^2$.
From these considerations we infer that the relaxation dynamics of the weights is in general a superposition of exponentials, dominated by the smallest eigenvalue $\omega_k$.
A more detailed investigation of the noise-driven time evolution is reserved for a future work.

\section{Summary and Outlook}

In the present study, we have analytically and numerically investigated the fluctuating dynamics of linear networks trained with mean-square loss.
The considered single-layer linear network trained with SGD solves an ordinary least-squares regression problem.
Besides being of theoretical interest, such linear models approximate the stationary training regime of deep nonlinear networks (see \cref{app_nonlinearNN}) and provide a basis for the construction of SGD-based Bayesian-sampling methods (see \cref{app_Bayesian}). 
Two-layer linear networks share important properties with deep nonlinear nets, such as nonlinear training dynamics, a non-convex loss landscape, and a sensitivity to initial conditions \cite{saxe_exact_2013,laurent_deep_2018,arora_implicit_2019}.

In our analytical calculations, we have considered uncorrelated Gaussian i.i.d.\ training data with targets generated by linearly transforming the inputs and adding noise (including the limit of fully random targets). 
Correlated or non-Gaussian input data will in general affect the gradient noise covariance [see \cref{eq_noise_cov_largeN,eq_noise_cov_Hess}] and thereby the weight statistics. 
However, for sufficiently large mini-batch sizes and small learning rates, the additive Gaussian approximation for the noise is usually warranted (see discussion in \cref{sec_noise_covar} and Refs.\ \cite{mandt_stochastic_2017, jastrzebski_three_2018, wu_revisiting_2021, wu_noisy_2019}).
Crucially, these assumptions enable a linearization of the nonlinear SGD dynamics of two-layer networks and thereby facilitate analytical calculations.
Our theory applies to neural networks in a (quasi-)stationary regime, characterized by a loss fluctuating around a non-zero mean value.
Typically, such a state is reached at the end of training of an underparameterized network.
Overparameterized networks, by contrast, can memorize their training data \cite{zhang_understanding_2017a} and approach a global optimum in which gradient noise vanishes \cite{sankararaman_impact_2019, du_gradient_2019a, du_gradient_2019, ma_power_2018, li_what_2021, mignacco_effective_2022}.
In this case, the loss does not become strictly stationary, but decays on a timescale much larger than the weight fluctuations. Indeed, the inverse-variance flatness relation (IVFR) has been considered in such a situation in Ref.\ \cite{feng_inverse_2021}.
Our perturbative approach to a two-layer network approximates such a quasi-stationary state by assuming that the mean of the weights evolve sufficiently slowly.

The main motivation of our present work was to study the simplest possible model that gives rise to the IVFR and to gain further insights into the nature of the SGD noise and the induced weight fluctuations. We have obtained the following main results:
\begin{enumerate}
	\item We have shown that one source of anisotropy of the weight fluctuations is the broken detailed balance of SGD \cite{chaudhari_stochastic_2017, kunin_limiting_2021}, which emerges here in the presence of label noise for mildly underparameterized nets. In the infinite sample limit, detailed balance is restored and the noise covariance becomes equal to the Hessian of the loss. 
	\item For two-layer linear nets, another source of fluctuation anisotropy stems from the rank deficiency of the drift and diffusion matrices that describe the stochastic gradient flow of the weight fluctuations in each layer. We expect this general consequence of the network structure to also hold for deeper linear networks. 
	\item Based on these insights, we have provided an analytical derivation of the IVFR for a two-layer linear network in the underparameterized regime. Our approach is based on a perturbative treatment of weight fluctuations around a quasi-stationary state and thereby adds a further perspective to existing explanations of the IVFR \cite{feng_inverse_2021, yang_stochastic_2023, adhikari_machine_2023}.  
	\item As a byproduct of our investigation, we have described the layer-wise continuum dynamics of SGD in two-layer linear nets, extending previous works that considered the deterministic gradient flow \cite{saxe_exact_2013, lampinen_analytic_2019, tarmoun_understanding_2021, braun_exact_2022}.
\end{enumerate}

Understanding weight fluctuations in neural networks is not only interesting from a theoretical perspective \cite{engel_statistical_2001, bahri_statistical_2020, huang_statistical_2021}, but has a number of practical implications. 
Noise-induced transition events, for instance, can be relevant for optimization in a rough loss landscape \cite{zhu_anisotropic_2018,daneshmand_escaping_2018,nguyen_first_2019,mori_powerlaw_2022}.
Recently, the fluctuation-dissipation ratio, which has its origin in statistical physics \cite{cugliandolo_effective_2011, marconi_fluctuationdissipation_2008}, has been introduced as a novel training measure \cite{yaida_fluctuationdissipation_2019}. 
In the context of differential privacy \cite{abadi_deep_2016}, random noise is added to the gradient updates.
In Ref.\ \cite{feng_inverse_2021}, the IVFR has been observed empirically for various deep nonlinear network architectures trained on image data under cross-entropy loss and explained within a phenomenological model of a random loss landscape.
Our derivation is based on layerwise perturbation of the SGD dynamics, which should also be applicable to deep nonlinear network models.
It will be interesting in the future to connect these two approaches and further clarify the relevance of the effects identified here (inter-layer coupling and rank deficiency) for fluctuation and loss anisotropy.
We have noted that, in the case of a two-layer linear network, the weight spectrum is significantly determined by the initialization and the IVFR emerges while the SGD dynamics is still transient despite a stationary loss. 
These aspects deserve a further investigation, which could consider also effects of non-Gaussianity and higher-order correlations stemming from the data or from nonlinearities in the dynamics \cite{simsekli_tailindex_2019, refinetti_neural_2022}.

\appendix

\section{Nonlinear networks}
\label{app_nonlinearNN}
Nonlinear networks can be linearized in the stationary state of training, provided deviations from the reference state are sufficiently small, which requires large batch sizes and small learning rates (see the discussion in \cref{sec_noise_covar}). 
Notably, for large widths, nonlinear networks reduce to linear models \cite{jacot_neural_2018,lee_wide_2019} and the associated limiting dynamics of SGD is a topic of recent interest \cite{sirignano_mean_2020, debortoli_quantitative_2020, gess_stochastic_2023, li_what_2021, ma_power_2018}.
For the linear models considered in the present study, steady-state weight fluctuations vanish in the overparameterized regime [see \cref{eq_zero_noise_OP}]. 

We denote the output function of the nonlinear network as $f(\xv,\wv)$, where $\wv\in\reals^{N\st{tot}}$ collectively represents the network parameters. 
Taylor expansion around a reference point $\wv^*$ renders 
\beq f(\xv,\wv)\simeq f(\xv,\wv^*) + (\wv-\wv^*)\cdot \nabla_\wv f(\xv,\wv^*) 
= b(\xv) + \wv\cdot \nabla_\wv f(\xv,\wv^*),
\eeq
with a bias term $b(\xv) = f(\xv,\wv^*) - \wv^*\cdot \nabla_\wv f(\xv,\wv^*)$ that is independent of $\wv$.
Defining a Jacobian matrix 
\beq \Phi=\begin{pmatrix}
	\nabla_\wv^T f(\xv_1,\wv^*) \\
	\vdots \\
	\nabla_\wv^T f(\xv^P,\wv^*)
\end{pmatrix} \in \reals^{P\times N},
\eeq 
we can write the MSE loss as 
\beq L(\wv) = \frac{1}{2P} \sum_\mu [f(\xv^\mu,\wv) - y^\mu]^2 = \frac{1}{2P}|| \wv^T \Phi - Y||^2 + \const.
\eeq 
up to a constant independent of $\wv$.
Accordingly, the linear framework of \cref{sec_linreg} can be adopted by replacing the design matrix $X$ by the Jacobian matrix $\Phi$ [see \cref{eq_LR_loss}].
We remark that the result resembles a random feature model, although the features $\pd_{w_i} f(\xv,\wv^*)$ are not independent \cite{cao_generalization_2019a,wei_more_2022, mei_generalization_2021, montanari_interpolation_2022}.

\section{Connection to Bayesian methods}
\label{app_Bayesian}

Since Bayesian neural networks involve the weight distribution as a central ingredient, it is informative to briefly review this approach and its connection to SGD-based sampling here \cite{wilson_bayesian_2020, jospin_handson_2022, roberts_principles_2022}.
The prediction $y^*$ for a new data point $x^*$ is computed from the likelihood $p(y^*|x^*,w)$ weighted by the posterior distribution $p(w|\Dcal)$: $p(y^*|x^*,\Dcal) = \int_w p(y^*|x^*,w) p(w|\Dcal) \d w$, where $\Dcal=\{x^\mu,y^\mu\}_\mu$ denotes the training data. According to Bayes' rule, the posterior is given by $p(w|\Dcal)\propto p(y|x,w)p(w)$, where $p(w)$ is the prior distribution of the weights and $p(y|x,w)\propto \exp(- \Lcal(x,y,w))$ is the likelihood function represented here in terms of a loss $\Lcal$. A regularized loss can accordingly be introduced by writing $p(w|\Dcal)\propto \exp(-\Lcal\st{reg}(x,y,w))$ with $\Lcal\st{reg} = \Lcal + \chi ||w||^2_2$, where the last equation holds for a Gaussian prior $p(w)$, which gives rise to an $L_2$-regularization. 
Classical (non-stochastic) training amounts to approximating the posterior as $p(w|\Dcal) = \delta(w-w\st{map})$ using the point estimate $w\st{map}=\text{argmin}_w \mathbb{E}_\Dcal\Lcal\st{reg}(x,y,w)$, where $\mathbb{E}_\Dcal$ denotes the average over the training data. 

In the stationary state of SGD, weight samples are drawn from \cref{eq_wts_cov,eq_wts_pdf}, which represents a different distribution from the true posterior. Thus, in order to use SGD as a Bayesian sampler, several methods have been proposed in the literature, such as preconditioning \cite{li_preconditioned_2015,mandt_stochastic_2017} or injecting Gaussian white noise into the SGD dynamics \cite{welling_bayesian_2011, ma_complete_2015,adhikari_machine_2023}. This isotropic white noise is constructed such that it dominates over the minibatch-sampling noise in the limit of small learning rates, implying $C+Q\to\Imat$ in \cref{eq_wts_cov}.

\section{Covariance of mini-batches}
\label{app_cov_MB}

Consider (possibly vectorial) quantities
\beq G_B = \frac{1}{S} \sum_{\mu\in\Bcal} g^\mu,\qquad G = \bra G_B\ket_\Bcal = \frac{1}{P}\sum_{\mu=1}^P g^\mu,
\eeq  
where $G_B$ is defined as an average over mini-batches $\Bcal$ of size $S=|\Bcal|$, while $G$ is given by the average of $G_B$ over the full batch of size $P>S$.
The covariance of $G_B$ and mini-batch sampling with replacement can be calculated, following similar steps as in Appendix E of \cite{ziyin_strength_2021}:
\beq\begin{split}
\mathrm{cov}(G_B) &\equiv \bra (G_B-\bra G_B\ket_\Bcal) (G_B-\bra G_B\ket_\Bcal)^T\ket_\Bcal = \bra G_B G_B^T\ket_\Bcal - G G^T \\
&= \frac{1}{S} \left[ \frac{1}{P} \sum_{\mu=1}^P g^\mu (g^\mu)^T - G G^T\right].
\end{split}\eeq 
The corresponding expression for sampling without replacement takes the same form but with the prefactor $(P-S)/(S(P-1))$ instead of $1/S$ \cite{ziyin_strength_2021}. The two cases coincide if $S=1$ or $P\gg S$.

\section{Additive noise approximation}
\label{app_addnoise}
We justify in the following the use of the additive noise approximation in the stationary training regime of a linear network.
We start from the Langevin \cref{eq_SGD_Langevin}, which we write 
\beq \dot \wv = -H(\wv-\wv^*)+\sqrt{\eta} \boldsymbol\zeta,\qquad \eta\equiv \frac{\lambda}{S}
\label{app_SGD_Lang}\eeq 
with $\wv^*$ given by \cref{eq_LR_sol}.
The correlations of the noise $\boldsymbol\zeta$ are described by the large-sample number limit [\cref{eq_noise_cov_largeN}]:
\beq \bra \boldsymbol\zeta(t) \boldsymbol\zeta(t')^T\ket = S C(\wv)\delta(t-t') = \left[ H \bar \wv \bar \wv^T H + \tr(H \bar \wv \bar \wv^T)H + \sigma_\epsilon^2 H \right]\delta(t-t'),
\label{app_SGD_noise}\eeq 
which is sufficient for the purpose of estimating the importance of the various contributions.
We consider $\eta$ as a smallness parameter and assume an expansion of the form 
\beq \wv=\wv_0 + \sqrt{\eta} \wv_1 + \Ocal(\eta).
\label{app_SGD_wexp}\eeq 
Analogously, we expand the noise as $\boldsymbol\zeta=\boldsymbol\zeta_0 + \boldsymbol\zeta_1+\ldots$, where the correlations of the new noises $\boldsymbol\zeta_i$ follow successively by evaluating \cref{app_SGD_noise} with \cref{app_SGD_wexp}.

Inserting \cref{app_SGD_wexp} into \cref{app_SGD_Lang} readily yields 
\beq \wv_0=\wv^*.
\eeq 
Next, using $\wv^* = \uv + H^{-1}J$ [\cref{eq_LR_sol}] renders 
\beq \bra \boldsymbol\zeta_0(t) \boldsymbol\zeta_0(t')^T\ket = C_0 \delta(t-t'), \qquad C_0 = [ J J^T  + \tr(J J^T H^{-1})H + \sigma_\epsilon^2 H ] \sim \sigma_x^2 \sigma_\epsilon^2 ,
\label{app_SGD_C0}\eeq 
and the Langevin equation at the order $\sqrt{\eta}$:
\beq \dot \wv_1 = -H \wv_1 + \boldsymbol\zeta_0.
\eeq 
The variance of the resulting stationary solution is given by [cf.\ \cref{eq_wts_cov_detbal}]
\beq \bra \wv_1 \wv_1^T\ket = \onehalf H^{-1} C_0 \sim \sigma_\epsilon^2
\eeq 
Using this result in the expansion of \cref{app_SGD_noise} provides the $\Ocal(\eta)$-term of the noise correlation:
\beq \bra \boldsymbol\zeta_1(t) \boldsymbol\zeta_1(t')^T\ket = C_1 \delta(t-t'),\qquad 
C_1 = \frac{\eta}{2} \left[ H \bra w_1 w_1^T\ket H + \tr(H \bra w_1 w_1^T\ket) H \right]  \sim \eta \sigma_x^4\sigma_\epsilon^2 
\label{app_SGD_C1}
\eeq
and a corresponding $\Ocal(\eta^{1/2})$-term with correlation $C_{1/2}=\sqrt{\eta/2}\, J \wv_1^T H + \ldots \sim \sqrt{\eta}\sigma_x^2 \sigma_\epsilon^2$ stemming from the cross terms between $\wv^*$ and $\wv_1$.
A comparison of the relative magnitude of these contributions and the one in \cref{app_SGD_C0} indicates that the multiplicative noise $\boldsymbol\zeta_1$ remains subdominant in the stationary state if $\lambda/S\ll 1$ and $\lambda \sigma_x^2/S\ll 1$ are fulfilled, which is the case for typical setups.
A more rigorous analysis based on adiabatic elimination techniques \cite{venturelli_tracer_2022, gardiner_stochastic_2009} for the Fokker-Planck equation [\cref{eq_FP}] is reserved for a future work.

\section{Noise covariance}
\label{app_noise_cov}
Assuming whitened inputs, \cref{eq_2L_noise_expr,eq_sigmaYX_squared} imply:
\begin{align}
\lambda^{-1}\cov(\Rcal_1)_{ij,kl} &= (W_{2}^{0,T} \Sigma_{0}^{yx})_{il} (\Sigma_{0}^{yx,T} W_{2}^{0} )_{jk} + \sigma_x^2 \bra y^2\ket \delta_{jl} (W_2^{0,T} W_2^0)_{ik}  - \sigma_x^2 (W_2^{0,T} \Sigma_0^{yx} w^{*,T} W_2^0)_{ik} \delta_{jl} \nonumber \\
&\qquad - \sigma_x^2 (W_2^{0,T} w^{*} \Sigma_0^{yx,T} W_2^0)_{ik} \delta_{jl} - \sigma_x^2 (W_2^{0,T} \Sigma_0^{yx})_{il} (w^{*,T} W_2^{0})_{jk} - \sigma_x^2 (W_2^{0,T} w^{*})_{il} (\Sigma_0^{yx,T} W_2^0)_{jk} \nonumber \\
&\qquad + \sigma_x^4 (W_2^{0,T} w^* w^{*,T} W_2^{0})_{ik} \delta_{jl} + \sigma_x^4 (W_2^{0,T} w^*)_{il} (w^{*,T} W_2^{0} )_{jk} \nonumber \\
&= \sigma_x^2 \bra y^2\ket \delta_{jl} (W_2^{0,T} W_2^0)_{ik} - (W_2^{0,T} \Sigma_0^{yx} \Sigma_0^{yx,T} W_2^0)_{ik} \delta_{jl} = \sigma_x^2 \bra y^2\ket \delta_{jl} \left( 1-\sfrac{1}{P}\right) (W_2^{0,T} W_2^0)_{ik} ,
\end{align}
\begin{align}
\lambda^{-1}\cov(\Rcal_2)_{ij,kl} &= (\Sigma_0^{yx} W_1^{0,T})_{il} (W_1^{0} \Sigma_0^{yx,T})_{jk} + \sigma_x^2 \bra y^2\ket \delta_{ik} (W_1^0 W_1^{0,T})_{jl}  - \sigma_x^2 (\Sigma_0^{yx} W_1^{0,T})_{il} (W_1^{0} w^{*,T})_{jk} \nonumber \\
&\qquad - \sigma_x^2 (\Sigma_0^{yx} w^{*,T})_{ik} (W_1^{0} W_1^{0,T})_{jl}  - \sigma_x^2 (w^* W_1^{0,T})_{il} (W_1^{0} \Sigma_0^{yx,T})_{jk} - \sigma_x^2 (w^{*}\Sigma_0^{yx,T})_{ik} (W_1^{0} W_1^{0,T})_{jl} \nonumber\\
&\qquad + \sigma_x^4 (w^* W_1^{0,T})_{il} (W_1^{0} w^{*,T})_{jk} + \sigma_x^4 (w^* w^{*,T})_{ik} (W_1^0 W_1^{0,T})_{jl} \nonumber\\
&=  \sigma_x^2  \bra y^2\ket \delta_{ik} \left(1-\sfrac{1}{P}\right) (W_1^0 W_1^{0,T})_{jl} ,
\end{align}
\begin{align}
\lambda^{-1}\cov(\Rcal_1,\Rcal_2)_{ij,kl} &= (W_2^{0,T}\Sigma_0^{yx} W_1^{0,T})_{il} \Sigma_{0,jk}^{yx,T} + \sigma_x^2\bra y^2\ket W^{0,T}_{2,ik} W^{0,T}_{1,jl}  -\sigma_x^2 (W_2^{0,T} \Sigma_0^{yx} w^{*,T})_{ik} W_{1,jl}^{0,T} \nonumber\\
&\qquad - \sigma_x^2 (W_2^{0,T} w^* \Sigma_0^{yx,T})_{ik} W_{1,jl}^{0,T}   -\sigma_x^2 (W_2^{0,T} \Sigma_0^{yx} W_1^{0,T})_{il} w^{*,T}_{jk} - \sigma_x^2 (W_2^{0,T} w^* W_1^{0,T})_{il} \Sigma_{0,jk}^{yx,T} \nonumber\\
&\qquad + \sigma_x^4 (W_2^{0,T} w^* w^{*,T})_{ik} W_{1,jl}^{0,T} + \sigma_x^4 (W_2^{0,T} w^* W_1^{0,T})_{il} w^{*,T}_{jk} \nonumber\\
&=  \sigma_x^2\bra y^2\ket W^{0,T}_{2,ik} W^{0,T}_{1,jl}  -  (W_2^{0,T} \Sigma_0^{yx} \Sigma_0^{yx,T})_{ik} W_{1,jl}^{0,T} =  \sigma_x^2\bra y^2\ket (1-1/P) W^{0,T}_{2,ik} W^{0,T}_{1,jl} ,
\end{align} 
which leads to \cref{eq_2L_noisecov}.



%

\end{document}